\documentclass[10pt,journal,compsoc]{IEEEtran}
\ifCLASSOPTIONcompsoc
  \usepackage[nocompress]{cite}
\else
  \usepackage{cite}
\fi
\ifCLASSINFOpdf
\else
\fi
\usepackage{textcomp}
\usepackage{multirow}
\usepackage{graphicx}
\usepackage[normalem]{ulem}
\useunder{\uline}{\ul}{}
\usepackage{subfigure}
\usepackage{url}
\usepackage{booktabs}
\usepackage{cite}
\usepackage{hyperref}
\usepackage{ifsym}
\usepackage{comment}
\hypersetup{
    colorlinks=true,
    linkcolor=red,
    citecolor=green,
}

\makeatletter
\IEEEtriggercmd{\reset@font\normalfont\fontsize{6.55pt}{6.55pt}\selectfont} 
\makeatother
\IEEEtriggeratref{1}

\begin{document}
%
\title{Human Action Recognition from Various Data Modalities: A Review}
%
%
%
%
\author{Zehua~Sun,~Qiuhong~Ke,~Hossein~Rahmani,~Mohammed~Bennamoun,~Gang~Wang,~and~Jun~Liu 
 \IEEEcompsocitemizethanks{
 \IEEEcompsocthanksitem Zehua Sun and Jun Liu are with Singapore University of Technology and Design, Singapore. (Corresponding author: Jun Liu)
\protect\\
E-mail: zehua.sun@my.cityu.edu.hk; jun\_liu@sutd.edu.sg.


\IEEEcompsocthanksitem Qiuhong Ke is with The University of Melbourne, Australia.
\protect\\
E-mail: qiuhong.ke@unimelb.edu.au.

\IEEEcompsocthanksitem Hossein Rahmani is with Lancaster University, UK.
\protect\\
E-mail: h.rahmani@lancaster.ac.uk.

\IEEEcompsocthanksitem Mohammed Bennamoun is with The University of Western Australia.
\protect\\
E-mail: mohammed.bennamoun@uwa.edu.au.

\IEEEcompsocthanksitem Gang Wang is with Alibaba Group, China.
\protect\\
E-mail: wanggang@ntu.edu.sg.

}

\thanks{
~~~~This work is supported by National Research Foundation, Singapore under its AI Singapore Programme (AISG Award No: AISG-100E-2020-065), and SUTD SRG. This work is also supported by TAILOR, a project funded by EU Horizon 2020 research and innovation programme under GA No 952215.
}

\thanks{
\textcolor{blue}{~~~~This manuscript has been accepted by IEEE Transactions on Pattern Analysis and Machine Intelligence (TPAMI) - DOI: \href{https://doi.org/10.1109/TPAMI.2022.3183112}{10.1109/TPAMI.2022.3183112}. This is the up-to-date version (updated in June 2022). We plan to update this arXiv version yearly to cover the latest advances in the field of human action recognition.}
}

}

%
%

\markboth{IEEE Transactions on Pattern Analysis and Machine Intelligence}%
{Human Action Recognition from Various Data Modalities: A Review}
\IEEEtitleabstractindextext{%
\begin{abstract}
Human Action Recognition (HAR) aims to understand human behavior and assign a label to each action. It has a wide range of applications, and therefore has been attracting increasing attention in the field of computer vision. 
Human actions can be represented using various data modalities, such as RGB, skeleton, depth, infrared, point cloud, event stream, audio, acceleration, radar, and WiFi signal, which encode different sources of useful yet distinct information and have various advantages depending on the application scenarios. Consequently, lots of existing works have attempted to investigate different types of approaches for HAR using various modalities. In this paper, we present a comprehensive survey of recent progress in deep learning methods for HAR based on the type of input data modality.
Specifically, we review the current mainstream deep learning methods for single data modalities and multiple data modalities, including the fusion-based and the co-learning-based frameworks. We also present comparative results on several benchmark datasets for HAR, together with insightful observations and inspiring future research directions.
\end{abstract}

}

\maketitle

\IEEEdisplaynontitleabstractindextext

%
\IEEEpeerreviewmaketitle

\renewcommand{\baselinestretch}{0.9} 

\IEEEraisesectionheading{\section{Introduction}}

%
%
%
%
\IEEEPARstart{H}{uman} Action Recognition (HAR), i.e., recognizing and understanding human actions, is crucial for a number of real-world applications. It can be used in visual surveillance systems \cite{lin2008human} to identify dangerous human activities, and autonomous navigation systems \cite{lu2020driver} to perceive human behaviors for safe operation. Besides, it is important for a number of other applications, e.g., video retrieval \cite{hu2007semantic}, human-robot interaction \cite{rodomagoulakis2016multimodal}, and entertainment \cite{shotton2011real}. 

In the early days, most of the works focused on using RGB or gray-scale videos as input for HAR \cite{poppe2010survey}, due to their popularity and easy access. Recent years have witnessed an emergence of works \cite{donahue2015long, liu2017skeleton, rahmani20163d, jiang2017learning, wang20203dv, ghosh2019spatiotemporal, liang2019audio, zeng2014convolutional, kim2015human, wang2019temporal} using other data modalities, such as skeleton, depth, infrared sequence, point cloud, event stream, audio, acceleration, radar, and WiFi for HAR. 
This is mainly due to the development of different kinds of accurate and affordable sensors, and the distinct advantages of different modalities for HAR depending on the application scenarios. 

Specifically, according to the visibility, data modalities can be roughly divided into two categories, namely, visual modalities and non-visual modalities. As shown in Table \ref{1}, RGB, skeleton, depth, infrared sequence, point cloud, and event stream are visually ``intuitive'' for representing human actions, and can be seen as visual modalities. Generally, visual modalities are very effective for HAR. Among them, RGB video data is the most common data type for HAR, which has been widely used in surveillance and monitoring systems. 
Skeleton data encodes the trajectories of human body joints. It is succinct and efficient for HAR when the action performing does not involve objects or scene context. 
Point cloud and depth data, which capture the 3D structure and distance information, are popularly used for HAR in robot navigation and self-driving applications. 
In addition, infrared data can be utilized for HAR in even dark environments, while the event stream keeps the foreground movement of the human subjects and avoids much visual redundancy, making it suitable for HAR as well. 

Meanwhile, audio, acceleration, radar, and WiFi, etc. are non-visual modalities, i.e., they are not visually ``intuitive'' for representing human behaviors. Nevertheless, these modalities can also be used for HAR in some scenarios which require the protection of privacy of subjects. 
Among them, audio data is suitable for locating actions in the temporal sequences, while acceleration data can be adopted for fine-grained HAR. 
Besides, as a non-visual modality, radar data can even be used for through-wall HAR. 
More details of different modalities are discussed in Section \ref{SINGLE MODALITY}. 

Single modality-based HAR has been extensively investigated in the past decades \cite{poppe2010survey, aggarwal2011human, ren2020survey, chen2013survey, slim2019survey, yousefi2017survey, li2019survey}.  
However, since different modalities have different strengths and limitations for HAR, the fusion of multiple data modalities and the transfer of knowledge across modalities to enhance the accuracy and robustness of HAR, have also received great attention recently \cite{cardenas2018multimodal, mahasseni2016regularizing}. 
More specifically, fusion means the combination of the information of two or more modalities to recognize actions. For example, audio data can serve as complementary information of the visual modalities to distinguish the actions ``putting a plate'' and ``putting a bag'' \cite{kazakos2019epic}. 
Besides fusion, some other methods have also exploited co-learning, i.e., transferring knowledge across different modalities to strengthen the robustness of HAR models. 
For example, in \cite{song2020modality}, the skeleton data serves as an auxiliary modality enabling the model to extract more discriminative features from RGB videos for HAR. 

Considering the significance of using different single modalities for HAR and also leveraging their complementary characteristics for fusion and co-learning-based HAR, we review existing HAR methods from the perspective of data modalities. 
Specifically, we review the mainstream deep learning architectures that use single data modalities for HAR, and also the methods taking advantage of multiple modalities for enhanced HAR. 
\textcolor{black}{
Note that in this survey, we mainly focus on HAR from short and trimmed video segments, i.e., each video contains only one action instance. 
}
We also report and discuss the benchmark datasets. 
The main contributions of this review are summarized as follows. 


   \noindent\textbf{(1)} To the best of our knowledge, this is the first survey paper that comprehensively reviews the HAR methods from the perspective of various data modalities, including RGB, depth, skeleton, infrared sequence, point cloud, event stream, audio, acceleration, radar, and WiFi.
   
   \noindent\textbf{(2)} We comprehensively review the multi-modality-based HAR methods, and categorize them into two types, namely, multi-modality fusion-based approaches and cross-modality co-learning-based approaches. 
   
   \noindent\textbf{(3)} We focus on reviewing the more recent and advanced deep learning methods for HAR, and hence provide the readers with the state-of-the-art approaches.
   
   \noindent\textbf{(4)} We provide comprehensive comparisons of existing methods and their performance on several benchmark datasets (e.g., Tables \ref{2}, \ref{3}, \ref{4}, \ref{5}), with brief summaries and insightful discussions. 

The rest of this paper is organized as follows.
Section \ref{SINGLE MODALITY} reviews HAR methods using different single data modalities. Section \ref{MULTI-MODALITY} introduces multi-modality HAR methods. The benchmark datasets are listed in Section \ref{DATASET}. Finally, potential future development of HAR is discussed in Section \ref{DISCUSSION}.

\section{Single Modality} \label{SINGLE MODALITY}
Different modalities can have different characteristics with corresponding advantages and disadvantages, as shown in Table \ref{1}. Hence numerous works \cite{donahue2015long, liu2017skeleton, rahmani20163d, jiang2017learning, wang20203dv, ghosh2019spatiotemporal, liang2019audio, zeng2014convolutional, kim2015human, wang2019temporal} exploited various modalities for HAR. Below we review the methods that use single data modalities, including RGB, skeleton, depth, infrared, point cloud, event stream, audio, acceleration, radar, and WiFi, for HAR. 

\begin{table}[t]
\caption{\footnotesize Action samples of different data modalities (with pros and cons). 
} 
\vspace{-3mm}
\centering
\resizebox{0.5\textwidth}{!}{
\setlength{\tabcolsep}{1.0pt}
\begin{tabular}{c|c|c|l|l}
\toprule[1pt]
\multicolumn{2}{c|}{\textbf{Modality}} &
  \textbf{Example} &
  \multicolumn{1}{c|}{\textbf{Pros}} &
  \multicolumn{1}{c}{\textbf{Cons}} \\ \hline
\multirow{6}{*}{\rotatebox{90}{\textbf{ Visual Modality }}} &
  RGB &
  \begin{tabular}[c]{@{}c@{}}\includegraphics[width=0.23\linewidth]{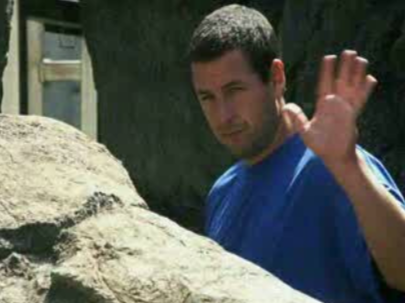}\\  Hand-waving \cite{kuehne2011hmdb}\end{tabular} &
  \begin{tabular}[c]{@{}l@{}}$\cdot$ Provide rich appearance \\ ~~information\\  \\ $\cdot$ Easy to obtain and operate\\  \\ $\cdot$ Wide range of applications\\ \end{tabular} &
  \begin{tabular}[c]{@{}l@{}}$\cdot$ Sensitive to viewpoint\\ \\$\cdot$ Sensitive to background\\ \\$\cdot$ Sensitive to illumination \\ 
  \end{tabular} \\ \cline{2-5} 
 &
  \begin{tabular}[c]{@{}c@{}}3D \\ Skeleton\end{tabular} &
  \begin{tabular}[c]{@{}c@{}}\includegraphics[width=0.23\linewidth]{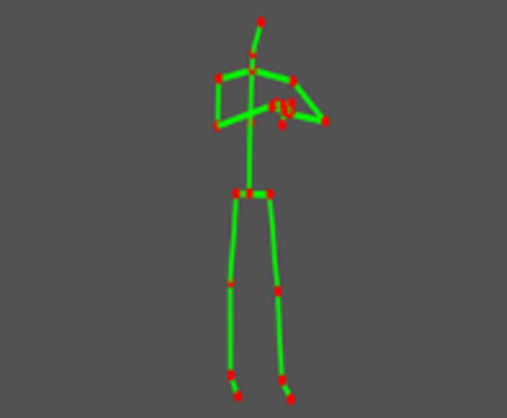}\\Looking at watch \cite{liu2017pku}\end{tabular} &
  \begin{tabular}[c]{@{}l@{}}$\cdot$ Provide 3D structural \\ ~~information of subject pose \\  \\ $\cdot$ Simple yet informative\\  \\ $\cdot$ Insensitive to viewpoint\\  \\ $\cdot$ Insensitive to background \\  \end{tabular} &
  \begin{tabular}[c]{@{}l@{}}$\cdot$ Lack of appearance \\~~information\\  \\ $\cdot$ Lack of detailed shape \\~~information \\ \\ $\cdot$ Noisy  \end{tabular} \\ \cline{2-5} 
 &
  Depth &
  \begin{tabular}[c]{@{}c@{}}\includegraphics[width=0.23\linewidth]{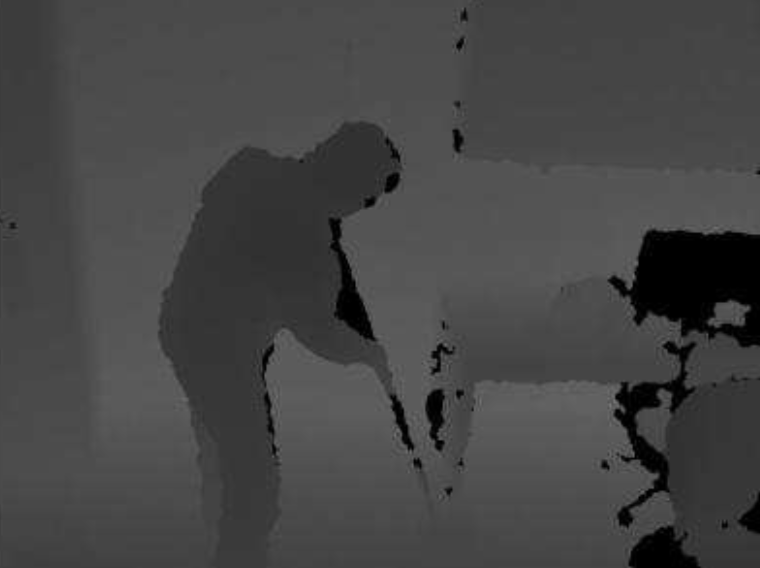}\\Mopping floor \cite{ni2011rgbd}\end{tabular} &
  \begin{tabular}[c]{@{}l@{}}$\cdot$ Provide 3D structural \\~~information \\  \\ $\cdot$ Provide geometric shape \\~~information \\  
  \end{tabular} &
  \begin{tabular}[c]{@{}l@{}}$\cdot$ Lack of color and texture \\~~information\\ \\ $\cdot$ Limited  workable \\~~distance\\ \end{tabular} \\ \cline{2-5} 
 &
  \begin{tabular}[c]{@{}c@{}} Infrared \\ Sequence \end{tabular} &
  \begin{tabular}[c]{@{}c@{}}\includegraphics[width=0.23\linewidth]{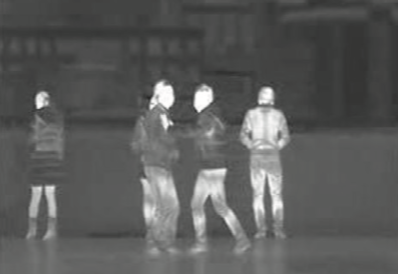}\\ Pushing \cite{gao2016infar}\end{tabular} &
  \begin{tabular}[c]{@{}l@{}}\begin{tabular}[c]{@{}l@{}}$\cdot$ Workable in dark \\~~environments \end{tabular}
  \end{tabular} 
  &
  \begin{tabular}[c]{@{}l@{}}$\cdot$ Lack of color and texture \\~~information\\ 
  \\ $\cdot$ Susceptible to sunlight \\  \end{tabular} \\ \cline{2-5} 
 &
  \begin{tabular}[c]{@{}c@{}} Point \\ Cloud \end{tabular} &
  \begin{tabular}[c]{@{}c@{}}\includegraphics[width=0.23\linewidth]{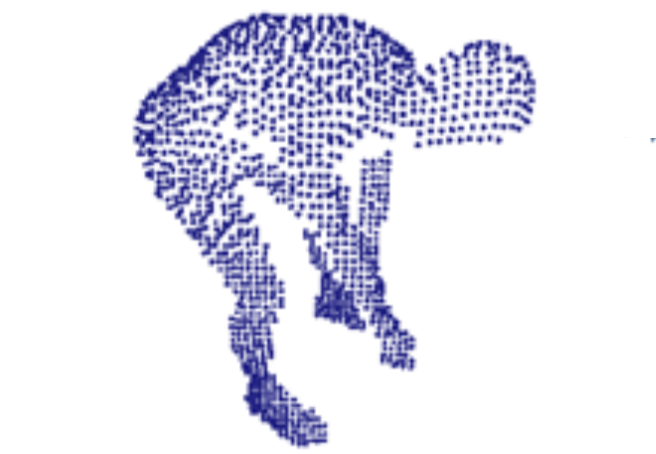}\\Bending over \cite{cheng2016orthogonal}\end{tabular} &
  \begin{tabular}[c]{@{}l@{}}$\cdot$ Provide 3D information\\  \\ $\cdot$ Provide geometric shape \\~~information\\  \\ $\cdot$ Insensitive to viewpoint 
  \end{tabular} 
  &
  \begin{tabular}[c]{@{}l@{}}$\cdot$ Lack of color and texture \\~~information\\  \\ $\cdot$ High computational \\~~complexity
  \end{tabular} \\
  \cline{2-5} 
  &
  \begin{tabular}[c]{@{}c@{}}Event \\Stream \end{tabular}&
  \begin{tabular}[c]{@{}c@{}}\includegraphics[width=0.23\linewidth]{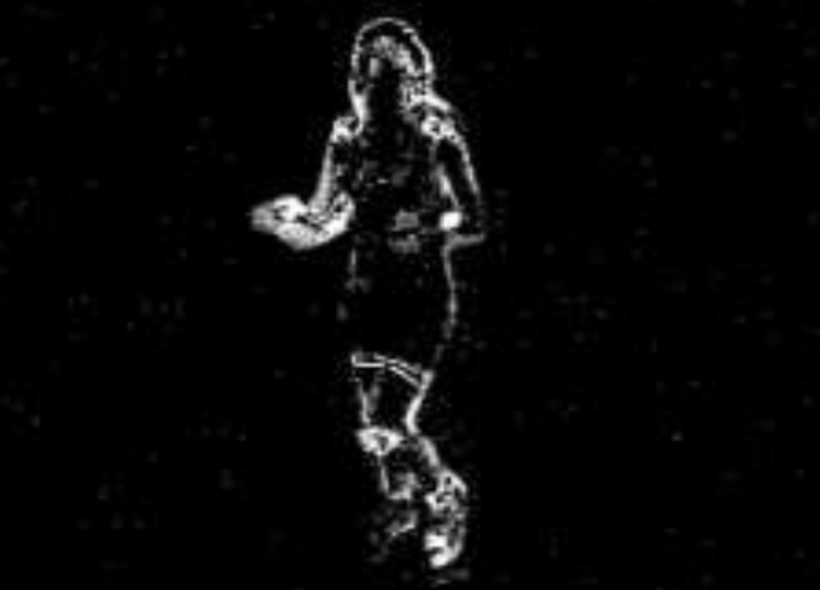}\\Running \cite{calabrese2019dhp19}\end{tabular} &
  \begin{tabular}[c]{@{}l@{}}$\cdot$ Avoid much visual \\ ~~redundancy\\  \\ $\cdot$ High dynamic range\\  \\ $\cdot$ No motion blur 
  \end{tabular} 
  &
  \begin{tabular}[c]{@{}l@{}}$\cdot$ Asynchronous output\\  \\ $\cdot$ Spatio-temporally sparse \\  \\ $\cdot$ Capturing device is \\~~relatively expensive \end{tabular} \\ \hline
\multirow{4}{*}{\rotatebox{90}{\textbf{Non-visual Modality}}} &
  Audio &
  \begin{tabular}[c]{@{}c@{}}\includegraphics[width=0.23\linewidth]{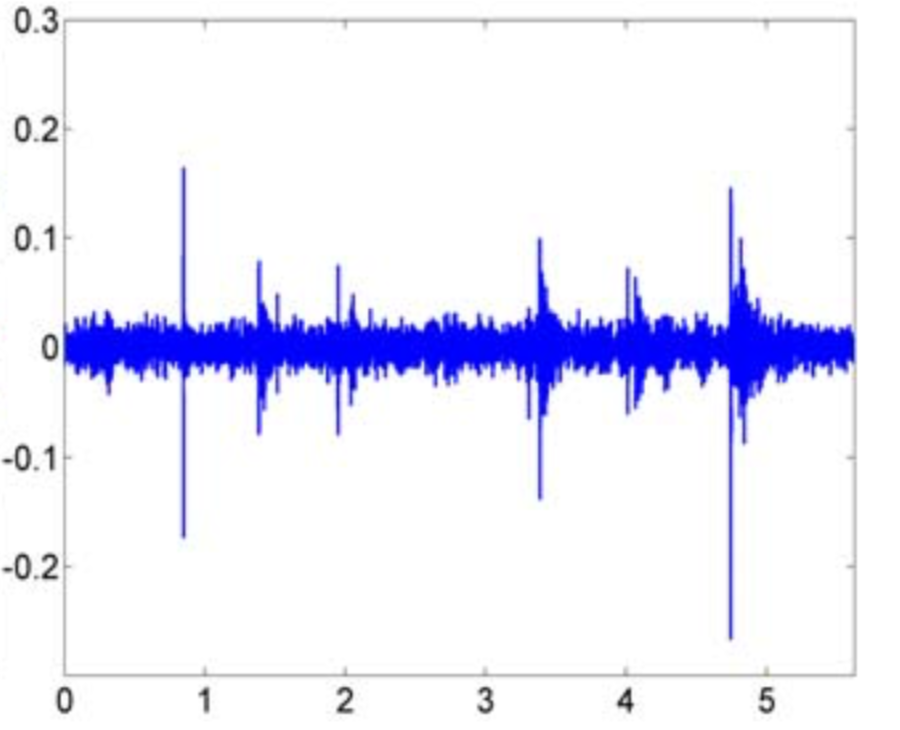}\\  Audio wave of jumping\cite{ofli2013berkeley}\end{tabular} &
  \begin{tabular}[c]{@{}l@{}} $\cdot$ Easy to locate actions in \\ ~~temporal sequence \\ \end{tabular} & 
  \begin{tabular}[c]{@{}l@{}}$\cdot$ Lack of appearance \\ ~~information \\ \end{tabular} \\ \cline{2-5} 
 &
  Acceleration &
  \begin{tabular}[c]{@{}c@{}}\includegraphics[width=0.23\linewidth]{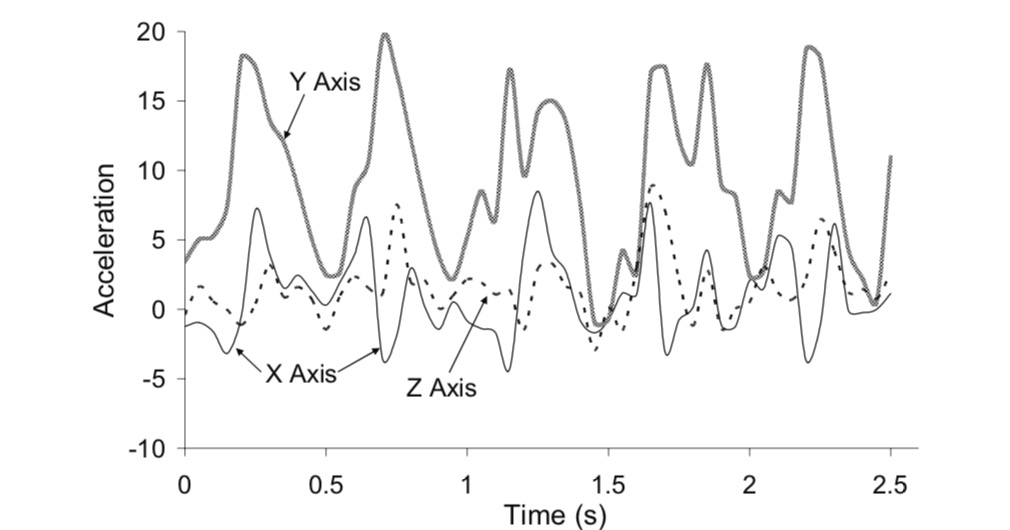}\\ Acceleration measurements \\of walking \cite{kwapisz2011activity}\end{tabular} &
  \begin{tabular}[c]{@{}l@{}}$\cdot$ Can be used for fine-grained \\~~HAR  \\  \\ $\cdot$ Privacy protecting\\ \\ $\cdot$ Low cost \\\end{tabular} &
  \begin{tabular}[c]{@{}l@{}}$\cdot$ Lack of appearance \\~~information \\ \\ $\cdot$ Capturing device needs \\~~to be carried by subject\\ \end{tabular} \\ \cline{2-5} 
 &
  Radar &
  \begin{tabular}[c]{@{}c@{}}\includegraphics[width=0.4\linewidth]{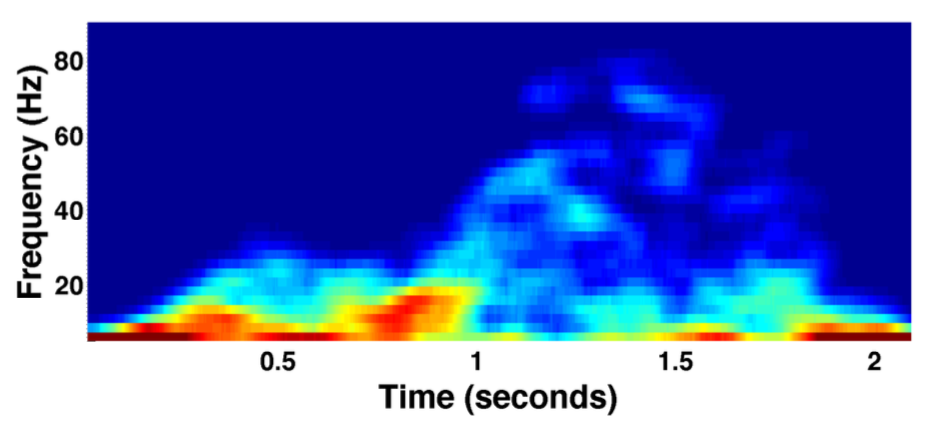}\\  Spectrogram 
  of falling \cite{wang2015understanding}\end{tabular} &
  \begin{tabular}[c]{@{}l@{}}$\cdot$ Can be used for through-wall \\~~HAR\\ \\ $\cdot$ Insensitive to illumination\\ \\ $\cdot$ Insensitive to weather\\ \\ $\cdot$ Privacy protecting \end{tabular} &
  \begin{tabular}[c]{@{}l@{}}$\cdot$ Lack of appearance \\~~information\\  \\ $\cdot$ Capturing device is \\~~relatively expensive
  \\  \end{tabular} \\ \cline{2-5} 
 &
  WiFi &
  \begin{tabular}[c]{@{}c@{}}\includegraphics[width=0.3\linewidth]{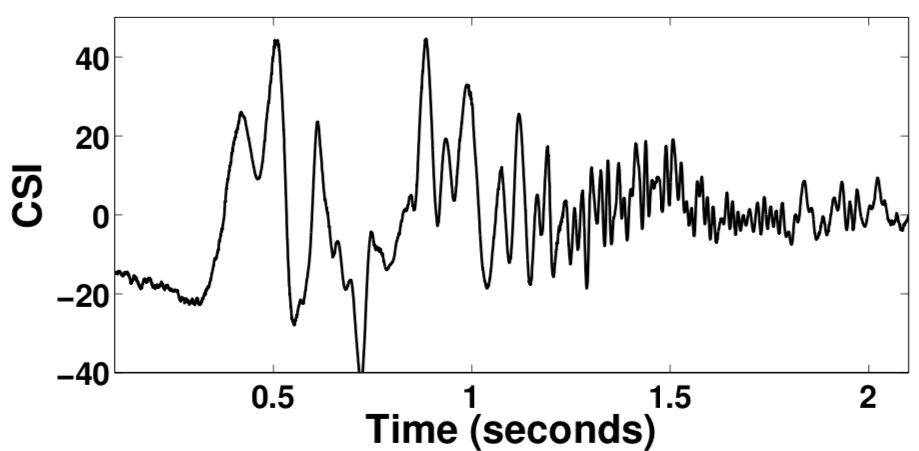}\\  CSI waveform of falling \cite{wang2015understanding}\end{tabular} &
  \begin{tabular}[c]{@{}l@{}} $\cdot$ Simple and convenient 
\\  \\ $\cdot$ Privacy protecting\\  \\ $\cdot$ Low cost \end{tabular} &
   \begin{tabular}[c]{@{}l@{}}$\cdot$ Lack of appearance \\~~information\\ \\ $\cdot$ Sensitive to environments \\ \\ $\cdot$ Noisy
\end{tabular} \\ 
\bottomrule[1pt]
\end{tabular} 
}
\label{1}
\end{table}

\subsection{RGB MODALITY} 
The RGB modality generally refers to images or videos (sequences of images) captured by RGB cameras which aim to recreate what human eyes see. 
RGB data is usually easy to collect, and it contains rich appearance information of the captured scene context. RGB-based HAR has a wide range of applications, such as visual surveillance \cite{lin2008human}, autonomous navigation \cite{lu2020driver}, and sport analysis \cite{soomro2014action}. 
However, action recognition from RGB data is often challenging, owing to the variations of backgrounds, viewpoints, scales of humans, and illumination conditions. Besides, RGB videos have generally large data sizes, leading to high computational costs when modeling the spatio-temporal context for HAR. 

In this subsection, we review the methods which use the RGB modality for HAR. 
Specifically, since videos contain the temporal dynamics of human motions that are often crucial for HAR, most of the existing works focused on using videos to handle HAR \cite{chaquet2013survey}, and only a few approaches utilized static images \cite{delaitre2010recognizing, yao2010grouplet, sharma2012discriminative}. 
Therefore, we focus, in the following, on reviewing RGB video-based HAR methods. 

In the pre-deep learning era, many hand-crafted feature-based approaches, including the space-time volume-based methods \cite{gorelick2007actions}, space-time interest point (STIP)-based methods \cite{laptev2005space}, and trajectory-based methods \cite{wang2011action}, were well-designed for RGB video-based HAR.
Recently, with the great progress of deep learning techniques, various deep learning architectures have also been proposed. 
Due to the strong representation capability and superior performance of deep learning-based methods, current mainstream research in this field focuses on designing different types of deep learning frameworks. Consequently, we 
review, in the following, the advanced deep learning works for RGB-based HAR, which can be mainly divided into four categories, namely, two-stream 2D Convolutional Neural Network (CNN), Recurrent Neural Network (RNN), 3D CNN, and Transformer-based methods, as shown in Table \ref{2}. 
In particular, Section \ref{sec:two-stream} mainly reviews the two-stream methods and their extensions
(e.g., multi-stream architectures), 
which use 2D CNNs as their backbone models. 
Section \ref{RNN-RGB} mainly reviews RGB-based HAR methods which mainly employ RNN models together with 2D CNNs as feature extractors.
In Section \ref{sec:3dcnn}, we review the 3D CNN-based methods and their extensions. 
Section \ref{sec:Transformer} introduces recent advances in Transformer-based methods for HAR. 

\subsubsection{Two-Stream 2D CNN-Based Methods} \label{sec:two-stream}
\begin{figure}[t]
\centering
\subfigure[Schema of two-stream 2D CNN-based methods. ]{
\begin{minipage}[t]{1\linewidth}
\centering
\includegraphics[height=0.8in]{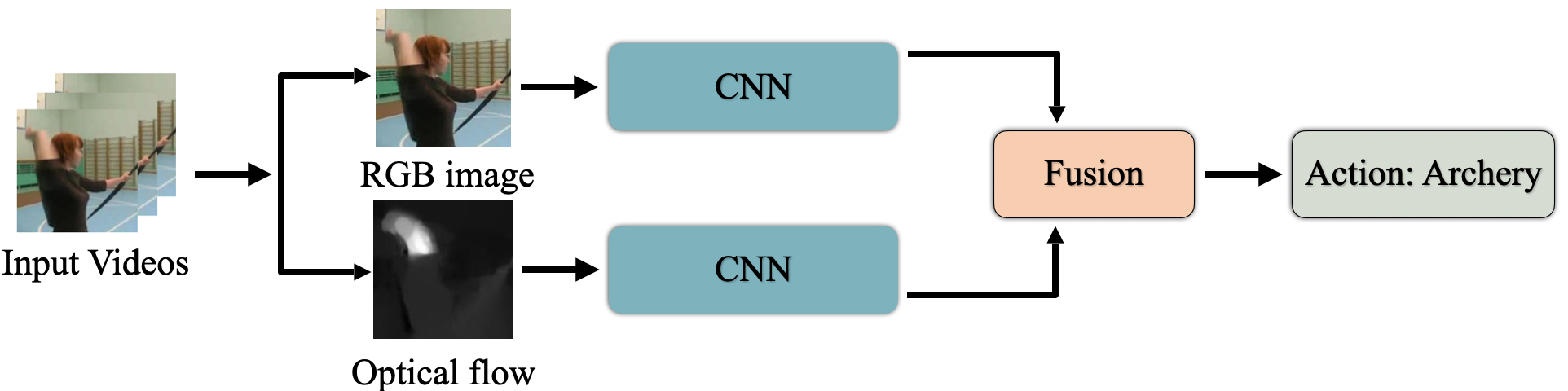}
\label{fig:TwoStream}
\end{minipage}%
}%

\subfigure[Schema of LSTM-based methods.]{
\begin{minipage}[t]{1\linewidth}
\centering
\includegraphics[height=0.6in]{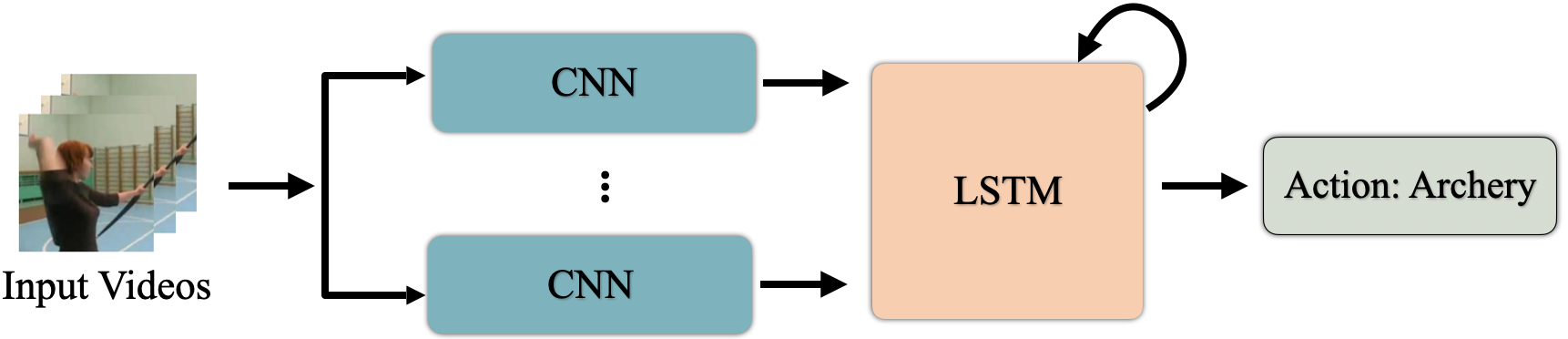}
\label{fig:LSTM}
\end{minipage}%
}%

\subfigure[Schema of 2D convolution.]{
\begin{minipage}[t]{0.5\linewidth}
\centering
\includegraphics[width=\textwidth]{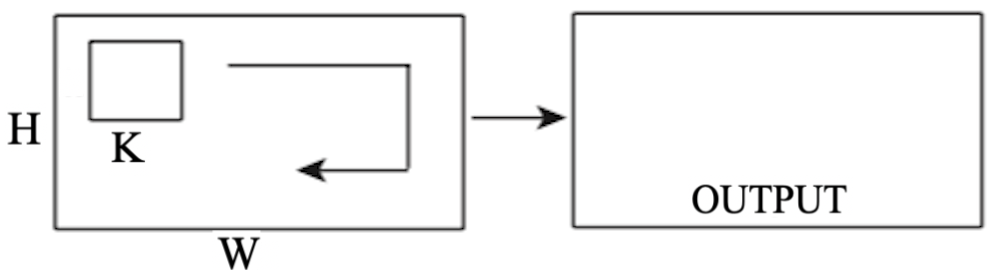}
\label{fig:tran2015learning1}
\end{minipage}%
}%
\subfigure[Schema of 3D convolution.]{
\begin{minipage}[t]{0.5\linewidth}
\centering
\includegraphics[width=\textwidth]{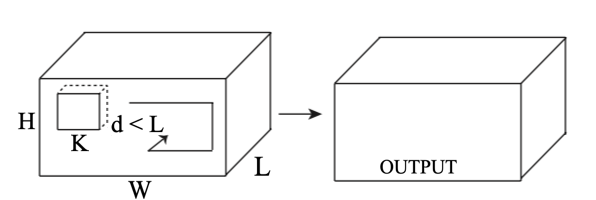}
\label{fig:tran2015learning2}
\end{minipage}
}%

\subfigure[Schema of Transformer-based methods.]{
\begin{minipage}[t]{1\linewidth}
\centering
\includegraphics[height=0.83in]{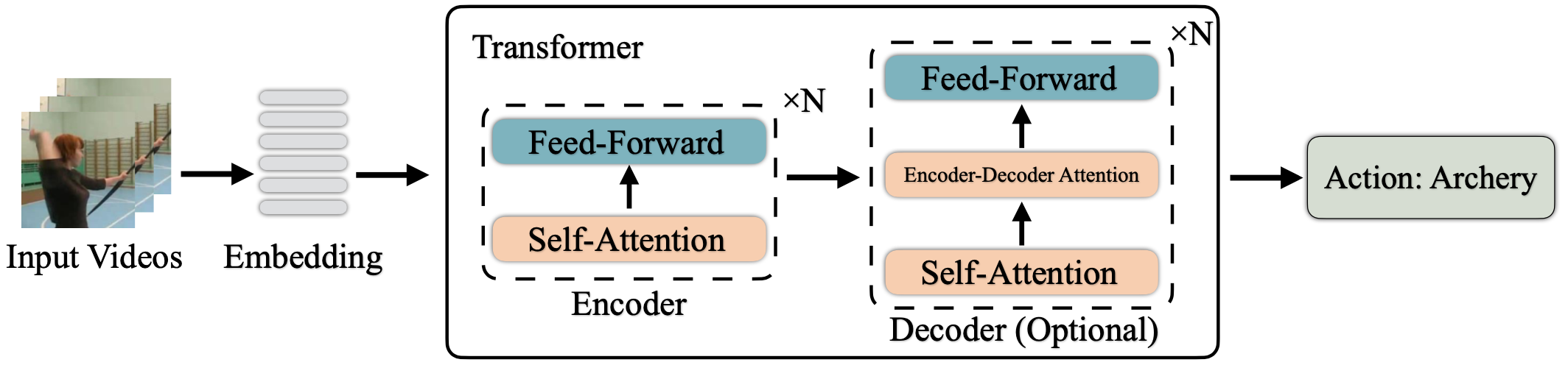}
\label{fig:Transformer}
\end{minipage}%
}%
\centering
\vspace{-3mm}
\caption{\footnotesize Illustration of RGB-based deep learning methods for HAR. 
(c) and (d) are originally shown in \cite{tran2015learning}.}
\end{figure}

As the name suggests, the two-stream 2D CNN framework generally contains two 2D CNN branches taking different input features extracted from the RGB videos for HAR, and the final result is usually obtained through fusion strategies, as shown in Fig. \ref{fig:TwoStream}. 
In this section, we review the classic two-stream methods \cite{simonyan2014two, karpathy2014large} 
and also their extensions \cite{bilen2017action}. 

As a classic two-stream framework, Simonyan and Zisserman \cite{simonyan2014two} proposed a two-stream CNN model consisting of a spatial network and a temporal network. 
More specifically, given a video, each individual RGB frame and multi-frame-based optical flows were fed to the spatial stream and temporal stream, respectively. Hence, appearance features and motion features were learned by these two streams for HAR. 
Finally, the classification scores of these two streams were fused to generate the final classification result. 
In another classic work, Karpathy et al. \cite{karpathy2014large} fed low-resolution RGB frames and high-resolution center crops 
to two separate streams to speed up the computation. Different fusion strategies were investigated to model the temporal dynamics in videos. Several studies endeavored to extend and improve over these classic two-stream CNNs, and they are reviewed below. 

To produce better video representations for HAR, Wang et al. \cite{wang2015action} fed multi-scale video frames and optical flows to a two-stream CNN to extract convolutional feature maps, which were then sampled over the spatio-temporal tubes centered at the extracted trajectories. Finally, the resulting features were aggregated using Fisher Vector representation \cite{sanchez2013image} followed by a SVM for HAR.
Ch{\'e}ron et al. \cite{cheron2015p} used the positions of human body joints to crop multiple human body parts from the RGB and optical flow images, which were passed through a two-stream network for feature extraction followed by a SVM classifier for HAR. 

There are also several other works \cite{wang2016temporal, girdhar2017actionvlad, diba2017deep, feichtenhofer2017spatiotemporal}, which extended the two-stream framework to extract long-term video-level information for HAR. 
Wang et al. \cite{wang2016temporal} divided each video into three segments and processed each segment with a two-stream network. The classification scores of the three segments were then fused by an average pooling method to produce the video-level prediction. 
Instead of fusing the scores of the segments as in \cite{wang2016temporal}, Diba et al. \cite{diba2017deep} aggregated the features of the segments by element-wise multiplication. 
Girdhar et al. \cite{girdhar2017actionvlad} extracted features from sampled appearance and motion frames based on a two-stream framework, and used the vocabulary of ``action words'' to aggregate the features into a single video-level representation for classification. 
Feichtenhofer et al. \cite{feichtenhofer2017spatiotemporal} multiplied the appearance residual features with the motion information at the feature level for gated modulation. 
\textcolor{black}{
Zong et al. \cite{zong2021motion} extended the two-stream CNN in \cite{feichtenhofer2017spatiotemporal} to a three-stream CNN by adding the motion saliency stream to better capture the salient motion information.
}
Bilen et al. \cite{bilen2017action} constructed dynamic images from RGB sequences and optical flow sequences to summarize the long-term global appearance and motion via rank pooling \cite{fernando2016rank}. The dynamic images together with the original RGB images and optical flow information were then fed to a multi-stream framework for HAR. Dynamic images have also been adopted by some other works as representations of videos for HAR \cite{wang2017structured, xiao2019action, wang2018cooperative}. 
Besides, the two-stream network has also been extended to a two-stream Siamese network to extract features from the frames before an action happens (precondition) and the frames after the action (effect), and the action was then represented as a transformation between the two sets of features \cite{wang2016actions}.

To tackle the high computational cost of computing accurate optical flow, some works \cite{zhang2016real} aimed to mimic the knowledge of the flow stream during training, in order to avoid the usage of optical flow during testing. 
Zhang et al. \cite{zhang2016real} proposed a teacher-student framework, which transfers the knowledge from the teacher network trained on optical flow data to the student network trained on motion vectors which can be obtained from compressed videos without extra calculation. Specifically, the knowledge was transferred by using the soft labels produced by the teacher model as an additional supervision to train the student network. 
Different from \cite{zhang2016real}, Piergiovanni and Ryoo \cite{Piergiovanni_2019_CVPR} proposed a trainable flow layer which captures the motion information without the need of computing optical flows. 

There have also been some works extending the two-stream CNN architectures in other aspects. 
Wang et al. \cite{wang2015towards} found that many two-stream CNNs were relatively shallow, and thus, they designed very deep two-stream CNNs for achieving better recognition results. 
Considering many frames of a video sequence could be irrelevant or useless for HAR, Kar et al. \cite{kar2017adascan} recursively predicted the discriminative importance of each frame for feature pooling. 
To perform HAR on low spatial resolution videos, Zhang et al. \cite{zhang2019two} proposed two video super-resolution methods producing high resolution videos, which were fed to the spatial and temporal streams to predict the action class. In the work of \cite{feichtenhofer2016convolutional}, several fusion strategies were studied, showing that it is effective to fuse the spatial and temporal networks at the last convolution layer hence reducing the number of parameters, yet keeping the accuracy. 

The two-stream 2D CNN architectures, that learn different types of information (e.g., spatial and temporal) from the input videos through separate networks and then perform fusion to get the final result, enable the traditional 2D CNNs to effectively handle the video data and achieve high HAR accuracy. 
However, this type of architectures is still not powerful enough for long-term dependency modeling, i.e., it has limitations in effectively modeling the video-level temporal information, while temporal sequence modeling networks, such as LSTM, can make up for it. 

\subsubsection{RNN-Based Methods} \label{RNN-RGB}
RNNs can be used to analyze temporal data due to the recurrent connections in their hidden layers. However, the traditional vanilla RNN suffers from the vanishing gradient problem, making it incapable of effectively modeling the long-term temporal dependency. 
Thus, most of the existing methods have adopted gated RNN architectures, such as Long-Short Term Memory (LSTM) \cite{du2017rpan, sun2017lattice, perrett2019ddlstm, meng2020ar}, to model the long-term temporal dynamics in video sequences. 


As shown in Fig. \ref{fig:LSTM}, RNN-based methods usually employ 2D CNNs, which serve as feature extractors, followed by an LSTM model for HAR. 
Donahue et al. \cite{donahue2015long} introduced the Long-term Recurrent Convolutional Network (LRCN), which consists of a 2D CNN to extract frame-level RGB features followed by LSTMs to generate a single action label.
Ng et al. \cite{yue2015beyond} extracted frame-level RGB and ``flow image'' features from pre-trained 2D CNNs, and then fed these features to a stacked LSTM framework for HAR. 
In the work of \cite{srivastava2015unsupervised}, an encoder LSTM was used to map an input video into a fixed-length representation, which was then decoded through a decoder LSTM to perform the tasks of video reconstruction and prediction in an unsupervised manner. 
Wu et al. \cite{wu2019liteeval} leveraged two LSTMs, which operate on coarse-scale and fine-scale CNN features cooperatively for efficient HAR. 
Majd and Safabakhsh \cite{majd2020correlational} proposed a C$^2$LSTM which incorporates convolution and cross-correlation operators to learn motion and spatial features while modeling temporal dependencies. 
Some other works \cite{ullah2017action, he2021db, zhao2020human} adopted the Bi-directional LSTM, 
which consists of two independent LSTMs to learn both the forward and backward temporal information, for HAR.


\textcolor{black}{
The introduction of attention mechanisms, in terms of spatial attention \cite{sharma2015action, ge2019attention, sudhakaran2019lsta, girdhar2017attentional}, temporal attention \cite{meng2019interpretable, wu2019adaframe}, and both spatial and temporal attention \cite{liu2020spatiotemporal, li2018videolstm}, have also benefited the LSTM-based frameworks to achieve better HAR performance. }
Sharma et al. \cite{sharma2015action} designed a multi-layer LSTM model, which recursively outputs attention maps weighting the input features of the next frame to focus on the important spatial features, leading to better recognition performance. 
Sudhakaran et al. \cite{sudhakaran2019lsta} introduced a recurrent unit with built-in spatial attention to spatially localize the discriminative information across a video sequence. 
LSTM has also been used to learn temporal attention to weight features of all frames \cite{meng2019interpretable}. 
\textcolor{black}{
Li et al. \cite{li2018videolstm} proposed a Video-LSTM, which incorporates convolutions and motion-based attention into the soft-attention LSTM \cite{ALSTM}, to better capture both spatial and motion information.}


Besides using LSTM, some works \cite{ballas2015delving, shi2017learning2, dwibedi2018temporal, kim2018discriminative} have used Gated Recurrent Units (GRUs) \cite{cho2014learning} for HAR, which can also alleviate the vanishing gradient problem of the vanilla RNN. 
Compared to LSTM, GRU has fewer gates, leading to fewer model parameters, yet it can usually provide a similar performance of LSTM for HAR \cite{chung2014empirical}. 

Several other works \cite{wu2015modeling, gammulle2017two, he2021db} focused on the hybrid architectures combining two-stream 2D CNN and RNN models for HAR. For example, Wu et al. \cite{wu2015modeling} utilized two-stream 2D CNN models to extract spatial and short-term motion features, which were fed to two LSTMs respectively to model longer-term temporal information for HAR. 

Generally, most of the RNN-based methods operate on the CNN features of frames instead of the high-dimensional frame images for HAR. 
Instead of
using 2D CNN, some other works \cite{baccouche2011sequential, wang2016beyond, shaheccv2020, wang2017two} used 3D CNNs as their feature extractors. In the following section, we review the 3D CNN-based methods.

\subsubsection{3D CNN-Based Methods} \label{sec:3dcnn}
Plenty of researches \cite{ji20123d, tran2015learning, zhang2020few, li2020directional} have extended 2D CNNs (Fig. \ref{fig:tran2015learning1}) to 3D structures (Fig. \ref{fig:tran2015learning2}), to simultaneously model the spatial and temporal context information in videos that is crucial for HAR. 
As one of the earliest works, Ji et al. \cite{ji20123d} segmented human subjects in videos by utilizing a human detector method, and then fed the segmented videos to a novel 3D CNN model to extract spatio-temporal features from videos. 
Unlike \cite{ji20123d}, Tran et al. \cite{tran2015learning} introduced a 3D CNN model, dubbed C3D, to learn the spatio-temporal features from raw videos in an end-to-end learning framework. 
However, these networks were mainly used for clip-level learning (e.g., 16 frames in each clip) instead of learning from full videos, which thus ignored the long-range spatio-temporal dependencies in videos.
Hence, several approaches  
focused on modeling the long-range spatio-temporal dependencies in videos. 
For example, Diba et al. \cite{diba2017temporal} extended DenseNet \cite{huang2017densely} with 3D filters and pooling kernels, and designed a Temporal 3D CNN (T3D), where the temporal transition layer can model variable temporal convolution kernel depths. T3D can densely and efficiently capture the appearance and temporal information at short, middle, and also long terms.
In their subsequent study \cite{diba2018spatio}, they introduced a new block embedded in some architectures such as ResNext and ResNet, which can model the inter-channel correlations of a 3D CNN with respect to the temporal and spatial features. 
Varol et al. \cite{varol2017long} proposed a Long-term Temporal Convolution (LTC) framework, which increases the temporal extents of 3D convolutional layers at the cost of reducing the spatial resolution, to model the long-term temporal structure. 
\textcolor{black}{
Hussein et al. \cite{hussein2019timeception} proposed multi-scale temporal-only convolutions, dubbed Timeception, to account for large variations and tolerate a variety of temporal extents in complex and long actions.} Wang et al. \cite{wang2018non} proposed a non-local operation, which models the correlations between any two positions in the feature maps to capture long-range dependencies. 
Besides, Li et al. \cite{li2020directional} proposed a Channel Independent Directional Convolution (CIDC) which can be attached to I3D \cite{carreira2017quo} to better capture the long-term temporal dynamics of the full input video. 

To enhance the HAR performance, 
several other works \cite{carreira2017quo,tran2018closer, zhou2018mict, zhu2018hidden,feichtenhofer2019slowfast} have investigated 3D CNN models with two-stream or multi-stream designs. 
For example, 
Carreira and Zisserman \cite{carreira2017quo} introduced the two-stream Inflated 3D CNN (I3D) inflating the convolutional and pooling kernels of a 2D CNN with an additional temporal dimension. 
Wang et al. \cite{wang2017two} integrated a two-stream 3D CNN with an LSTM model to capture the long-range temporal dependencies. 
Feichtenhofer et al. \cite{feichtenhofer2019slowfast} 
designed a two-stream 3D CNN framework containing a slow pathway and a fast pathway that operate on RGB frames at low and high frame rates to capture semantic and motion, respectively. 
Inspired by \cite{feichtenhofer2016spatiotemporal}, the features of the fast pathway were fused (by summation or concatenation) with the features of the slow pathway at each layer. 
Li et al. \cite{li2020spatio} introduced a two-stream spatio-temporal deformable 3D CNN with attention mechanisms to capture the long-range temporal and long-distance spatial dependencies.
Besides, deep learning frameworks combining 2D CNNs and 3D CNNs have also been investigated for HAR \cite{zolfaghari2018eco, wang2018appearance, Martinez_2019_ICCV}. For example, the ECO architecture \cite{zolfaghari2018eco} uses 2D CNNs to extract spatial features, which are stacked and then fed to 3D CNNs to model long-term dependencies for HAR.

Some works focused on addressing certain other problems in 3D CNNs. 
For example, Zhou et al. \cite{zhou2020spatiotemporal} analyzed the spatio-temporal fusion in 3D CNN from a probabilistic perspective. 
Yang et al. \cite{yang2020temporal} proposed a generic feature-level Temporal Pyramid Network (TPN) to model speed variations in performing actions. 
Kim et al. \cite{kim2020regularization} proposed a Random Mean Scaling (RMS) regularization method to address the problem of overfitting. 
\textcolor{black}{
The viewpoint variation problem has also been investigated \cite{piergiovanni2021recognizing, varol2021synthetic}.
Inspired by \cite{rahmani2016synthetic}, Varol et al. \cite{varol2021synthetic} proposed to train 3D CNN on synthetic action videos to address the problem of variations in viewpoints. 
Piergiovanni and Ryoo \cite{piergiovanni2021recognizing} proposed a geometric convolutional layer to learn view-invariant representations of actions by imposing the latent action representations to follow 3D geometric transformations and projections. 
}
Knowledge distillation
has also been investigated to improve motion representations of 3D CNN frameworks. 
For example, Stroud et al. \cite{stroud2020d3d} introduced a Distilled 3D Network (D3D) consisting of a student network and a teacher network, where the student network was trained on RGB videos and it also distilled knowledge from the teacher network that was trained on optical flow sequences. 
Crasto et al. \cite{crasto2019mars} transferred the knowledge of a teacher model trained on the optical flows to a 3D CNN student network operating on RGB videos, by minimizing the mean squared error between the feature maps of the two streams. 
\textcolor{black}{
Following the research line of handling the large computational cost of obtaining optical flow, 
Shou et al. \cite{shou2019dmc} proposed an adversarial framework with a lightweight generator to approximate flow information by refining the noisy and coarse motion vector available in the compressed video.
Differently, Wang et al. \cite{wang2020video} adopted an efficient learnable correlation operator to better learn motion information from 3D appearance features. 
Fayyaz et al. \cite{fayyaz20213d} addressed the problem of dynamically adapting
the temporal feature resolution within the 3D CNNs to reduce their computational cost. A Similarity Guided Sampling (SGS) module was proposed to enable 3D CNNs to dynamically adapt their computational resources by selecting the most informative and distinctive temporal features.
}


The 3D CNN-based methods are very powerful in modeling discriminative features from both the spatial and temporal dimensions for HAR. However, many 3D CNN-based frameworks contain a large number of parameters, and thus require a large amount of training data. 
\textcolor{black}{Therefore, some studies \cite{sun2015human, tran2018closer, qiu2017learning, Xie2018RethinkingSF, li2021ct} aimed to factorize 3D convolutions. 
}
Sun et al. \cite{sun2015human} proposed a Factorized spatio-temporal CNN (F$_{st}$CN), factorizing the 3D convolution to 2D spatial convolutional layers followed by 1D temporal convolutional layers.  Similarly, Qiu et al. \cite{qiu2017learning} decomposed the 3D convolution into a 2D convolution for the spatial domain, followed by a 1D convolution for the temporal domain, to economically and effectively simulate 3D convolutions.  Xie et al. \cite{Xie2018RethinkingSF} explored the performance of several variants of I3D \cite{carreira2017quo} by utilizing a combination of 3D and 2D convolutional filters in the I3D network, and introduced temporally separable convolutions and spatio-temporal feature gating to enhance HAR. Yang et al. \cite{yang2019asymmetric} proposed efficient asymmetric one-directional 3D convolutions to approximate the traditional 3D convolution. 
Besides, some other works \cite{lin2019tsm, sudhakaran2020gate, wang2021tdn, wang2021action} also utilized 2D CNNs with powerful temporal modules to decrease the computational cost of 3D CNNs. 
Lin et al. \cite{lin2019tsm} proposed a Temporal Shift Module (TSM), which shifts a part of the channels along the temporal dimension to perform temporal interaction between the features from adjacent frames. 
\textcolor{black}{ 
Unlike parameter-free temporal shift operations in \cite{lin2019tsm}, Sudhakaran et al. \cite{sudhakaran2020gate} introduced a lightweight Gate-Shift Module (GSM), which uses learnable spatial gating blocks for spatial-temporal decomposition of 3D convolutions. 
Wang et al. \cite{wang2021tdn} designed a two-level Temporal Difference Module (TDM) to capture both finer local and long-range global motion information. 
Wang et al. \cite{wang2021action} introduced an ACTION module leveraging multipath excitation to respectively model spatial-temporal, channel-wise, and motion patterns. 
}

The 3D CNN-based HAR methods generally perform spatio-temporal processing over limited intervals via the window-based 3D convolutional operations, where each convolutional operation only attends to relatively short-term context in videos. Meanwhile, RNN-based methods process video sequence elements recurrently and thus cannot model relatively long-term spatio-temporal dependency. However, Transformers can directly attend to complete video sequences via its scalable self-attention mechanism, and thus can effectively learn long-range spatio-temporal relationships in videos. Hence, many recent works have also investigated Transformer-based HAR in RGB videos. In the following section, we review the Transformer-based methods.

\begin{table}[t]
\caption{\footnotesize \textcolor{black}{Performance comparison of RGB video-based deep learning methods for HAR on the UCF101, HMDB51, and Kinectis-400 datasets. Note in RGB video-based HAR methods, some other information, e.g., optical flow or motion vector, may also be used along with the RGB data as input for HAR. For simplicity, `\%' after the value is omitted. `-' indicates the result is unavailable. `Flow' denotes optical flow.}}
\vspace{-3mm}
\centering
\resizebox{0.5\textwidth}{!}{
\setlength{\tabcolsep}{2.0pt}
\begin{tabular}{clccccc}
\toprule[1pt]
\multicolumn{2}{c}{\multirow{2}{*}{\textbf{Method}}}                                                & \multirow{2}{*}{\textbf{Year}} &\multirow{2}{*}{\textbf{Input}}& \multicolumn{3}{c}{\textbf{Dataset}}                      \\ \cline{5-7} 
\multicolumn{2}{c}{}                                                                                      &                             &                                & \textbf{UCF101} & \textbf{HMDB51} & \textbf{Kinetics-400} \\ \hline 

\multicolumn{1}{c|}{\multirow{16}{*}{\rotatebox{90}{\textbf{Two-stream/Multi-stream 2D CNN}}}} & Multiresolution CNN                   \cite{karpathy2014large}                                & 2014 & High-,Low-Resolution RGB                        & 65.4            & -                & -                     \\ 
\multicolumn{1}{c|}{}                                       & Two-stream CNN+SVM                  \cite{simonyan2014two}                                  & 2014 & RGB,Flow                          & 88.0            & 59.4             & -                     \\ 
\multicolumn{1}{c|}{}                                       & P-CNN                                 \cite{cheron2015p}                                      & 2015 & RGB,Flow                         & -               & -                & -                     \\ 
\multicolumn{1}{c|}{}                                       & Very deep two-stream                  \cite{wang2015towards}                                  & 2015 & RGB,Flow                          & 91.4            & -                & -                     \\ 
\multicolumn{1}{c|}{}                                       & TDD+iDT                                   \cite{wang2015action}                                   & 2015 & RGB,Flow                         & 91.5            & 65.9             & -                     \\ 
\multicolumn{1}{c|}{}                                       & EMV+RGB-CNN                         \cite{zhang2016real}                                    & 2016 & RGB,Motion Vector                          & 86.4            & -                & -                     \\ 
\multicolumn{1}{c|}{}                                       & Siamese Network                       \cite{wang2016actions}                                  & 2016 & RGB,Flow                          & 92.4            & 62.0             & -                     \\ 
\multicolumn{1}{c|}{}                                       & Feichtenhofer et al.                 \cite{feichtenhofer2016convolutional}                   & 2016 & RGB,Flow                          & 93.5            & 69.2             & -                     \\ 
\multicolumn{1}{c|}{}                                       & TSN                                   \cite{wang2016temporal}                                 & 2016 & RGB,Flow                          & 94.2            & 69.4             & -                     \\ 
\multicolumn{1}{c|}{}                                       & AdaScan+iDT                   \cite{kar2017adascan}                                   & 2017 & RGB,Flow                           & 91.3            & 61.0             & -                     \\ 
\multicolumn{1}{c|}{}                                       & ActionVLAD+iDT                      \cite{girdhar2017actionvlad}                            & 2017 & RGB,Flow                          & 93.6            & 69.8             & -                     \\ 
\multicolumn{1}{c|}{}                                       & Feichtenhofer et al.                 \cite{feichtenhofer2017spatiotemporal}                  & 2017 & RGB,Flow                          & 94.9            & 72.2             & -                     \\ 
\multicolumn{1}{c|}{}                                       & TLE: Bilinear                  \cite{diba2017deep}                                     & 2017 & RGB,Flow                          & 95.6            & 71.1             & -                     \\ 
\multicolumn{1}{c|}{}                                       & Bilen et al.                  \cite{bilen2017action}                                     & 2017 & RGB,Flow,Dynamic Images                            & 96.0            & 74.9             & -                     \\ 
\multicolumn{1}{c|}{}                                       & TSN+SoSR+ToSR                     \cite{zhang2019two}                                     & 2019 & RGB,Flow                          & 92.1            & 68.3             & -                     \\ 
\multicolumn{1}{c|}{}                                       & \textcolor{black}{MSM-ResNets}                    \cite{zong2021motion}                                     & 2021 & RGB,Flow,Saliency Map                         & 93.5            & 66.7             & -                     \\ \hline
\multicolumn{1}{c|}{\multirow{15}{*}{\rotatebox{90}{\textbf{RNN}}}}           
& Soft Attention Model                  \cite{sharma2015action}                                 & 2015 & RGB                           & -               & 41.3             & -                     \\ 
\multicolumn{1}{c|}{}                                       & LRCNs                                 \cite{donahue2015long}                                  & 2015 & RGB,Flow                           & 82.7 
& -                & -                     \\ 
\multicolumn{1}{c|}{}                                       & Composite LSTM Model                    \cite{srivastava2015unsupervised}                       & 2015 & RGB,Flow                          & 84.3            & -             & -                     \\ 
\multicolumn{1}{c|}{}                                       & Ng et al.                           \cite{yue2015beyond}                                    & 2015 & RGB,Flow                           & 88.6            & -                & -                     \\ 
\multicolumn{1}{c|}{} & Wu et al.                            \cite{wu2015modeling}                                   & 2015 &RGB,Flow                           & 91.3            & -                & -                     \\ 
\multicolumn{1}{c|}{}                                       & RNN-FV+iDT                                \cite{lev2016rnn}                                   & 2016 &RGB 
& 94.1            & 67.7             & -                     \\ 
 
\multicolumn{1}{c|}{}                                       & L$^2$STM                              \cite{sun2017lattice}                                   & 2017 & RGB,Flow                           & 93.6            & 66.2             & -                     \\ 
\multicolumn{1}{c|}{}                                       & shuttleNet                              \cite{shi2017learning2}                                   & 2017 & RGB,Flow                          & 95.4            & 71.7             & -                     \\ 
\multicolumn{1}{c|}{}                                       & \textcolor{black}{iDT(FV)+Objects+VideoLSTM}                              \cite{li2018videolstm}                                   & 2018 & RGB,Flow                          &  92.2           &64.9              & -                     \\ 
\multicolumn{1}{c|}{}                                       & CNN+TR-LSTM                         \cite{pan2019compressing}                               & 2019 & RGB                          & -               & 63.8             & -                     \\ 
\multicolumn{1}{c|}{}                                       & Ge et al.                            \cite{ge2019attention}                                  & 2019 & RGB,Flow                          & 92.8            & 67.1             & -                     \\ 
\multicolumn{1}{c|}{}                                       & AR-Net                         \cite{meng2020ar}                               & 2020 & RGB                           & -               & -             & -                     \\ 
\multicolumn{1}{c|}{}                                       & STS-ALSTM                             \cite{liu2020spatiotemporal}                            & 2020 & RGB,Flow,Saliency Map                          & 92.7            & 64.4             & -                     \\ 
\multicolumn{1}{c|}{}                                       & C$^2$LSTM                             \cite{majd2020correlational}                            & 2020 & RGB                           & 92.8            & 61.3             & -                     \\ 
\multicolumn{1}{c|}{}                                       & DB-LSTM                         \cite{he2021db}                               & 2021 & RGB,Flow                          & 97.3            & 81.2                & -                     \\ \hline 
\multicolumn{1}{c|}{\multirow{45}{*}{\rotatebox{90}{\textbf{3D CNN}}}}         & Ji et al.                            \cite{ji20123d}                                         & 2012 & RGB,Gradient,Flow                         & -               & -                & -                     \\ 
\multicolumn{1}{c|}{}                                       & F$_{st}$CN                            \cite{sun2015human}                                     & 2015 & RGB                           & 88.1            & 59.1             & -                     \\ 
\multicolumn{1}{c|}{}                                       & C3D+iDT+SVM                       \cite{tran2015learning}                                 & 2015 & RGB                        & 90.4            & -                & -                     \\ 
\multicolumn{1}{c|}{}                                       & scLSTM                                \cite{wang2016beyond}                                   & 2016 & RGB                          & 84.0            & 55.1             & -                     \\ 
\multicolumn{1}{c|}{}                                       & ST-ResNet$^*$+iDT                   \cite{feichtenhofer2016spatiotemporal}                  & 2016 & RGB,Flow                          & 94.6            & 70.3             & -                     \\ 
\multicolumn{1}{c|}{} & STPP                             \cite{wang2017two}                                      & 2017 & RGB,Flow                          & 92.6            & 70.5             & -                     \\ 
\multicolumn{1}{c|}{}& LTC$_{Flow+RGB}$+iDT                             \cite{varol2017long}                                    & 2017 & RGB,Flow                          & 92.7            & 67.2             & -                     \\ 
 \multicolumn{1}{c|}{}& T3D+TSN                               \cite{diba2017temporal}                                 & 2017 & RGB                         & 93.2            & 63.5             &  62.2 (T3D)                    \\ 
\multicolumn{1}{c|}{}                                       & P3D ResNet+iDT                      \cite{qiu2017learning}                                  & 2017 & RGB                          & 93.7            & -                & -                     \\ 
\multicolumn{1}{c|}{}&  Two-Stream I3D \cite{carreira2017quo}                    & 2017 & RGB,Flow                          & 97.9              & 80.2                & - \\
\multicolumn{1}{c|}{}                                       & NL I3D                       \cite{wang2018non}                               & 2018 & RGB                         & -            & -             & 77.7                  \\ 
\multicolumn{1}{c|}{}                                       & ARTNet with TSN                       \cite{wang2018appearance}                               & 2018 &  RGB                         & 94.3            & 70.9             & 80.0                  \\ 
\multicolumn{1}{c|}{}                                       & Two-stream MiCT-Net                   \cite{zhou2018mict}                                     & 2018 & RGB,Flow                          & 94.7            & 70.5             & -                     \\ 
\multicolumn{1}{c|}{}                                       & ECO\_{En}                             \cite{zolfaghari2018eco}                                & 2018 &  RGB                         & 94.8            & 72.4             & 70.0                  \\ 
\multicolumn{1}{c|}{}                                       & STC-ResNext 101 (64 frames)           \cite{diba2018spatio}                                   & 2018 & RGB                          & 96.5            & 74.9             &  68.7 (32 frames)     \\ 
\multicolumn{1}{c|}{}                                       & S3D-G           \cite{Xie2018RethinkingSF}                                   & 2018 &  RGB                         & 96.8            & 75.9             & 74.7       \\ 
\multicolumn{1}{c|}{}                                       & Hidden Two-stream CNN+I3D           \cite{zhu2018hidden}                                    & 2018 & RGB                           & 97.1            & 78.7             & -                     \\ 
\multicolumn{1}{c|}{}                                       & R(2+1)D-TwoStream                     \cite{tran2018closer}                                   & 2018 & RGB,Flow                           & 97.3            & 78.7             & 75.4                  \\ 
\multicolumn{1}{c|}{}                                       &         Martinez et al.            \cite{Martinez_2019_ICCV}                               & 2019 & RGB                           & -            & -             & 78.8                     \\ 
\multicolumn{1}{c|}{}                                       & SlowFast 16$\times$8, R101+NL                              \cite{feichtenhofer2019slowfast}                        & 2019 & RGB                           & -               & -                & 79.8                  \\ 
\multicolumn{1}{c|}{}                                       & irCSN-152                            \cite{ghadiyaram2019large}                        & 2019 &RGB                           & -               & -                & 82.8                 \\ 
\multicolumn{1}{c|}{}                                       & Asymmetric 3D-CNN+iDT                     \cite{yang2019asymmetric}                               & 2019 & RGB,RGBF                          & 92.6            & 65.4             & -                     \\ 
\multicolumn{1}{c|}{}                                       & \textcolor{black}{I3D RGB+DMC-Net (I3D)}                 \cite{shou2019dmc}                                     & 2019 & RGB,Residual,Motion Vector                           & 96.5            & 77.8             & -                     \\ 
\multicolumn{1}{c|}{}                                       & MARS                            \cite{crasto2019mars}                        & 2019 & RGB,Flow                          & 98.1               & 80.9                & 74.9                 \\ 
\multicolumn{1}{c|}{}                                       & SlowFast+RMS                          \cite{kim2020regularization}                               & 2020 & RGB                           & -            & -             & 76.3                     \\ 
\multicolumn{1}{c|}{}                                       & TPN-R101                          \cite{yang2020temporal}                               & 2020 &  RGB                         & -            & -             & 78.9                     \\ 
\multicolumn{1}{c|}{}                                       & X3D-XL                          \cite{feichtenhofer2020x3d}                               & 2020 &  RGB                         & -            & -             & 79.1                     \\
\multicolumn{1}{c|}{}                                       & \textcolor{black}{CorrNet-101}                          \cite{wang2020video}                               & 2020 &  RGB                         & -            & -             & 81.0                     \\ 
\multicolumn{1}{c|}{}                                       & STDA-ResNeXt-101                      \cite{li2020spatio}                                     & 2020 & RGB                          & 95.5            & 72.7             & -                     \\ 
\multicolumn{1}{c|}{}                                       & Zhou et al.                     \cite{zhou2020spatiotemporal}                                     & 2020 & RGB                          & 96.5            & -             & 72.5                     \\ 
\multicolumn{1}{c|}{}                                       & D3D                          \cite{stroud2020d3d}                                    & 2020 & RGB                          & 97.0            & 78.7             & 75.9                     \\ 
\multicolumn{1}{c|}{}                                       & CIDC (I3D-NL)                           \cite{li2020directional}                                    & 2020 & RGB                          & 97.9            & 75.2             & 75.6                     \\ 
\multicolumn{1}{c|}{}                                       & OmniSource (RGB \& Flow)                     \cite{duan2020omni}                                     & 2020 & RGB,Flow                          & 98.6            & 83.8             & -                     \\
\multicolumn{1}{c|}{}                                       & OmniSource (RGB)                     \cite{duan2020omni}                                     & 2020 & RGB                          & -            & -             & 83.6                     \\
\multicolumn{1}{c|}{}                                       & R(2+1)D+BERT (64f)                      \cite{kalfaoglu2020late}                                     & 2020 & RGB                          & 98.7            & 85.1            & -                     \\
\multicolumn{1}{c|}{}                                       & \textcolor{black}{X3D XL+AFTR${^\beta}$}                      \cite{fayyaz20213d}                                     & 2021 & RGB                          & -            & -            & 79.3                     \\
\multicolumn{1}{c|} {}& \textcolor{black}{CT-Net\_{EN}}                     \cite{li2021ct}                                     & 2021 & RGB                          & -            & -            & 79.8                     \\
\multicolumn{1}{c|}{}                                       & \textcolor{black}{Yang et al.}                      \cite{yang2021beyond}                                     & 2021 & RGB                          & -            & -            & 81.4                     \\
\multicolumn{1}{c|}{}                                       & \textcolor{black}{MetaUVFS}                      \cite{patravali2021unsupervised}                                     & 2021 & RGB                          & 76.4            & 47.6            & -                     \\
\multicolumn{1}{c|}{}                                       & \textcolor{black}{Slow I3D 8$\times$8+SFC}                      \cite{liu2021selective}                                     & 2021 & RGB                          & 94.5            & -            & 74.6                     \\
\multicolumn{1}{c|}{}                                       & \textcolor{black}{3DResNet+ATFR}                      \cite{fayyaz20213d}                                     & 2021 & RGB                          & 97.9            & 76.7            & -                     \\ 
\multicolumn{1}{c|}{}                                       & \textcolor{black}{NUTA network I3D-101}                      \cite{li2022nuta}                                     & 2022 & RGB                          & -            & -            & 78.9                     \\ 
\multicolumn{1}{c|}{}                                       & \textcolor{black}{SLIC}                      \cite{khorasgani2022slic}                                     & 2022 & RGB,Flow                          & 83.2            & 56.2            & -                     \\ 
\multicolumn{1}{c|}{}                                       & \textcolor{black}{SIN(AVG)}                      \cite{shi2022shuffle}                                     & 2022 & RGB                          & 91.6            & 74.6            & -                     \\ 
\multicolumn{1}{c|}{}                                       & \textcolor{black}{Omi et al.}                      \cite{omi2022model}                                     & 2022 & RGB                          & 95.8            & 75.0            & 70.1                     \\ \hline
\multicolumn{1}{c|}{\multirow{20}{*}{\rotatebox{90}{\textbf{Transformer}}}}         &   \textcolor{black}{Girdhar et al.}                      \cite{girdhar2019video}                                     & 2019 & RGB                          & -            & -            & -                    \\ 
\multicolumn{1}{c|}{}                                       & \textcolor{black}{GroupFormer}\cite{li2021groupformer}                                       & 2021 &  RGB                        & -               & -                & -                     \\
\multicolumn{1}{c|}{}                                       & \textcolor{black}{ViT-B-VTN}\cite{neimark2021video}                                       & 2021 &  RGB                         & -               & -                & 79.8                     \\
\multicolumn{1}{c|}{}                                       & \textcolor{black}{En-VidTr-L}\cite{zhang2021vidtr}                                       & 2021 &  Pixel Patch                         & -               & -                & 80.5                     \\
\multicolumn{1}{c|}{}                                       & \textcolor{black}{TimeSformer-L}\cite{bertasius2021space}                                       & 2021 &  RGB                        & -               & -                & 80.7                     \\
\multicolumn{1}{c|}{}                                       & \textcolor{black}{X-ViT}\cite{bulat2021space}                                       & 2021 &  RGB                        & -               & -                & 80.7                     \\
\multicolumn{1}{c|}{}                                       & \textcolor{black}{Mformer-HR}\cite{patrick2021keeping}                                       & 2021 &  RGB                        & -               & -                & 81.1                     \\
\multicolumn{1}{c|}{}                                       & \textcolor{black}{MViT-B, 64$\times$3}\cite{fan2021multiscale}                                       & 2021 &  RGB                        & -               & -                & 81.2                     \\
\multicolumn{1}{c|}{}                                       & \textcolor{black}{ViViT-H/16$\times$2 (JFT)}\cite{arnab2021vivit}                                       & 2021 &  RGB                        & -               & -                & 84.9                     \\
\multicolumn{1}{c|}{}                                       & \textcolor{black}{TokenLearner 16at18+L/10}\cite{ryoo2021tokenlearner}                                       & 2021 &  Image-Like Tensor                         & -               & -                & 85.4                     \\
\multicolumn{1}{c|}{}                                       & \textcolor{black}{SCT-L}\cite{zha2021shifted}                                       & 2021 &  RGB                        & 98.7               & 84.6                & 83.0                     \\
\multicolumn{1}{c|}{}                                       & \textcolor{black}{DETR}\cite{zhang2022efficient}                                       & 2022 &  RGB                        & -               & -                & -                     \\
\multicolumn{1}{c|}{}                                       & \textcolor{black}{UGPT}\cite{guo2022uncertainty}                                       & 2022 &  RGB                        & -               & -                & -                     \\
\multicolumn{1}{c|}{}                                       & \textcolor{black}{RegionViT-M}\cite{chen2022regionvit}                                       & 2022 &  RGB                        & -               & -                & 77.6                    \\
\multicolumn{1}{c|}{}                                       & \textcolor{black}{RViT-XL, 64$\times$3$\times$3}\cite{yang2022recurring}                                       & 2022 &  RGB                        & -               & -                & 81.5                    \\
\multicolumn{1}{c|}{}                                       & \textcolor{black}{DirecFormer}                      \cite{truong2022direcformer}                                     & 2022 & RGB                          & -            & -            & 82.8                     \\ 
\multicolumn{1}{c|}{}                                       & \textcolor{black}{UniFormer-B}\cite{li2022uniformer}                                       & 2022 &  RGB                        & -               & -                & 83.0                     \\
\multicolumn{1}{c|}{}                                       & \textcolor{black}{SIFAR-L-12}\cite{fan2021can}                                       & 2022 &  RGB                        & -               & -                & 84.2                     \\
\multicolumn{1}{c|}{}                                       & \textcolor{black}{MTV-H (WTS)}\cite{yan2022multiview}                                       & 2022 &  RGB                        & -               & -                & 89.1                     \\
\multicolumn{1}{c|}{}                                       & \textcolor{black}{SVT}\cite{ranasinghe2022self}                                       & 2022 &  RGB                        & 93.7               & 67.2                & 78.1                     \\

\bottomrule[1pt]
\end{tabular}%
}
\label{2}
\end{table}

\subsubsection{Transformer-Based Methods} \label{sec:Transformer}
Transformer \cite{vaswani2017attention} is a novel deep learning model leading the machine learning field recently, due to its strong capabilities and promising prospects. 
As shown in Fig.~\ref{fig:Transformer}, Transformer is composed of an encoder and a decoder. The encoder mainly consists of several self-attention blocks to encode the input sequence. The decoder shares a similar architecture as the encoder except an extra encoder-decoder attention mechanism in each block. 
This design enables Transformer to perform well concerning long-term dependency modeling, multi-modal fusion, and multi-task processing \cite{han2022survey, khan2021survey}.

Inspired by the success of Transformers in Natural Language Processing (NLP), plenty of works \cite{li2022uniformer, ryoo2021tokenlearner} employed Transformers for HAR from videos.  
Girdhar et al. \cite{girdhar2019video} proposed an Action Transformer network consisting of a Faster-RCNN~\cite{ren2015faster} style model followed by a Transformer for action localization and recognition. Particularly, the Faster-RCNN~\cite{ren2015faster} style network composes of an I3D model~\cite{carreira2017quo} as the base to generate initial spatio-temporal features and a Region Proposal Network to obtain object proposals for action localization. The Transformer treats the feature map of each particular subject as the query and features from the neighboring frames as the key and value. 
Bertasius et al. \cite{bertasius2021space} extended ViT \cite{dosovitskiy2020image} to videos by decomposing each video into a sequence of frame-level patches. Then, a divided attention mechanism was proposed to separately apply spatial and temporal attentions within each block of the model. 
Arnab et al. \cite{arnab2021vivit} proposed pure-Transformer network variants by factorizing different components of the Transformer encoder along the spatial and temporal dimensions. 
Truong et al. \cite{truong2022direcformer} proposed a spatio-temporal directed attention architecture (DirecFormer), which learns the right order of frames within the action video by exploiting the direction of attention in addition to the magnitude of attention between frames. 
Patrick et al. \cite{patrick2021keeping} injected a trajectory attention block into the Transformer network to enhance the robustness of HAR in dynamic scenes. The trajectory attention block first generates a set of trajectory tokens in the spatial dimension, and then performs pooling operation over the temporal dimension to aggregate information along the motion trajectories. 
Fan et al. \cite{fan2021multiscale} proposed a multi-scale pyramid of features, which hierarchically expands the channel capacity while reducing the spatial resolution in videos via a multi-head pooling attention. 
Yan et al. \cite{yan2022multiview} introduced Multiview Transformers, consisting of multiple individual encoders, each of which is specialized for a single input representation. Lateral connections between individual encoders are employed to effectively fuse information from different representations of the input video. 
Guo et al. \cite{guo2022uncertainty} proposed an Uncertainty Guided Probabilistic Transformer (UGPT) by extending the standard Transformer to a probabilistic one. Particularly, UGPT treats the attention values as random variables and models their distribution as Gaussian distribution allowing the model to quantify the uncertainties of the prediction. Then, an uncertainty-guided training algorithm is utilized to separately train two models that respectively focus on low-uncertainty and high-uncertainty data. During the inference, these two models are dynamically combined to produce the action class. 
Ranasinghe et al. \cite{ranasinghe2022self} proposed a Self-supervised Video Transformer (SVT) that learns cross-view and motion dependency from video clips with varying spatial sizes and frame rates. 

Although video Transformers have acquired promising results, they also suffer from severe memory and computational overhead. Researchers thus have made plenty of efforts to decrease computation complexity \cite{zha2021shifted, neimark2021video, pan2021ia, bulat2021space, fan2021can, chen2022regionvit} and memory cost \cite{zha2021shifted, pan2021ia, chen2022regionvit, yang2022recurring, zhang2021vidtr}. 
Fan et al. \cite{fan2021can} presented an efficient Super Image for Action Recognition (SIFAR) approach, which transforms the 3D video frames into 2D super images as the input to the Transformer network. 
Zha et al. \cite{zha2021shifted} introduced a Shifted Chunk Transformer (SCT), which is composed of a frame encoder and a clip encoder. The frame encoder relies on the image chunk and the shifted multi-head self-attention modules to capture fine-grained intra-frame representation and inter-frame variances, respectively. Specially, each individual frame is split into several local patches, which are fed into a hierarchical image chunk Transformer employing Locality-Sensitive Hashing (LSH) to significantly reduce the memory and computational consumption of matrix product in self-attention. Besides, the clip encoder is used to capture long-term temporal dependency. 
Neimark et al. \cite{neimark2021video} proposed a Video Transformer Network (VTN) to achieve a trade-off between speed and accuracy. VTN mainly relies on a spatial backbone (e.g., a 2D network) for frame-level feature processing and a temporal attention-based encoder, i.e., the Longformer model \cite{beltagy2020longformer}, for temporal modeling. 
Pan et al. \cite{pan2021ia} proposed an Interpretability-Aware REDundancy REDuction (IA-RED$^{2}$) framework, which learns to find the uncorrelated redundant patches, and hierarchically and dynamically drop them, resulting in a considerable reduction of computational complexity. IA-RED$^{2}$ is model-agnostic, task-agnostic, and inherently interpretable showing promising results in HAR. 
A local-window temporal attention strategy \cite{bulat2021space} and a separable attention mechanism \cite{zhang2021vidtr} have also achieved a considerable shrinkage of computational complexity and memory overhead without loss of accuracy, respectively. 
Yang et al. \cite{yang2022recurring} proposed a Recurrent Vision Transformer (RViT), which can be applied on both fixed-length and variant-length video clips without requiring huge memory. Specifically, RViT employs an attention gate mechanism, which operates in a recurrent manner, i.e., the output and current hidden state are determined by the current input frame and previous hidden state. Such a hidden state can capture spatio-temporal feature representations. 
Chen et al. \cite{chen2022regionvit} proposed a Regional-to-local attention-based Vision Transformer (RegionViT), which first generates regional and local tokens from each frame, and then performs interaction between a regional token and the associated local ones via information exchanging. Thus the local tokens can focus on a local field with global information, allowing the network to achieve a better trade-off between accuracy and complexity. Finally, the divided-space-time attention in TimeSformer~\cite{bertasius2021space} is adopted to model temporal variation. 

Apart from the above-mentioned architectures, i.e., two-stream 2D CNN, RNN, 3D CNN, and Transformer, there are also some other frameworks designed for HAR using RGB videos, such as 
Convolutional Gated Restricted Boltzmann Machines \cite{taylor2010convolutional}, \textcolor{black}{Graph-based Modeling \cite{li2021representing, zhou2021graph}}, and 4D CNNs \cite{zhang2020v4d}. 

In general, RGB data is the most commonly used modality for HAR in real-world application scenarios, since it is easy to collect and contains rich information. Besides, RGB-based HAR methods can leverage large-scale web videos to pre-train models for better recognition performance \cite{duan2020omni, ghadiyaram2019large}. 
However, it often requires complex computations for feature extraction from RGB videos. 
Moreover, HAR methods based on the RGB modality are often sensitive to viewpoint variations and background clutters, etc. Hence, HAR with other modalities, such as 3D skeleton data, has also received great attention, and is therefore discussed in the subsequent sections.

\subsection{SKELETON MODALITY} \label{sec:skeleton}
Skeleton sequences encode the trajectories of human body joints, which characterize informative human motions. Therefore, skeleton data is also a suitable modality for HAR. 
The skeleton data can be acquired by applying pose estimation algorithms on RGB videos \cite{sun2019deep, gong2022meta} or depth maps \cite{shotton2011real, liu2019synthetic}. 
It can also be collected with motion capture systems. Generally, human pose estimation is sensitive to viewpoint variations. 
Meanwhile, motion capture systems that are insensitive to view and lighting can provide reliable skeleton data. However, in many application scenarios, it is not convenient to deploy motion capture systems. 
Thus many recent works on skeleton-based HAR used skeleton data obtained from depth maps \cite{liu2019ntu} or RGB videos \cite{yan2018spatial}.

There are many advantages of using skeleton data for HAR, due to its provided body structure and pose information, its essentially simple and informative representation, its scale invariance, and its robustness against variations of clothing textures and backgrounds. 
Due to these advantages and also the availability of accurate and low-cost depth sensors, skeleton-based HAR has attracted much attention in the research community recently. 

Early works focused on extracting hand-crafted spatial and temporal features from skeleton sequences for HAR. The hand-crafted feature-based methods can be roughly divided into joint-based \cite{wang2013learning} and body part-based \cite{vemulapalli2014human} methods, depending on the used feature extraction techniques. 
Due to the strong feature learning capability, deep learning has been widely used for skeleton-based HAR, and has become the mainstream research in this field. Consequently, we review, in the following, the deep learning methods, which can be mainly divided into four categories: RNN, CNN, Graph Neural Network (GNN) or Graph Convolutional Network (GCN), and Transformer-based methods, as shown in Fig. \ref{fig:skeleton}. Table \ref{3} shows the results achieved by different skeleton-based HAR methods on two benchmark datasets. 

\subsubsection{RNN-Based Methods}
As mentioned in Section \ref{RNN-RGB}, RNNs and their gated variants (e.g., LSTMs) are capable of learning the dynamic dependencies in sequential data. Hence, various methods \cite{du2015hierarchical, zhang2017geometric, zhang2018fusing, liu2018skeleton_TIP} have applied and adapted RNNs and LSTMs to effectively model the temporal context information within the skeleton sequences for HAR. 

In one of the classical methods, Du et al. \cite{du2015hierarchical} proposed an end-to-end hierarchical RNN, dividing the human skeleton into five body parts instead of inputting the skeleton in each frame as a whole. These five body parts were then separately fed to multiple bidirectional RNNs, whose output representations were hierarchically fused to generate high-level representations of the action. 
Differential RNN (dRNN) \cite{veeriah2015differential} learns salient spatio-temporal information by quantifying the change of the information gain caused by the salient motions between frames. The Derivative of States (DoS) was proposed inside the LSTM unit to act as a signal controlling the information flow into and out of the internal state over time. 
Zhu et al. \cite{zhu2016co} introduced a novel mechanism for the LSTM network to achieve automatic co-occurrence mining, since co-occurrence intrinsically characterizes the actions. 
Sharoudy et al. \cite{shahroudy2016ntu} proposed a Part-aware LSTM (P-LSTM) by introducing a mechanism to simulate the relations among different body parts inside the LSTM unit. 
Liu et al. \cite{liu2016spatio, liu2017skeleton} extended the RNN design to both the temporal and spatial domains. 
Specifically, they utilized the tree structure-based skeleton traversal method to further exploit the spatial information, and trust gates to deal with noise and occlusions. 
In their subsequent study \cite{liu2017global}, an attention-based LSTM network named Global Context-Aware Attention LSTM (GCA-LSTM) 
was proposed to selectively focus on the informative joints using the global context information. 
The GCA-LSTM network contains two LSTM layers, where the first layer encodes the skeleton sequence and outputs a global context memory, while the second layer outputs attention representations to refine the global context. Finally, a softmax classifier outputs the HAR result.
In the work of \cite{wang2017modeling}, a two-stream RNN structure was proposed to model both the temporal dynamics and spatial configurations. 
Deep LSTM with spatio-temporal attention was proposed in \cite{song2017end}, where a spatial attention sub-network and a temporal attention sub-network work jointly under the main LSTM network. 
To model variable temporal dynamics of skeleton sequences, Lee et al. \cite{lee2017ensemble} proposed an ensemble Temporal Sliding LSTM (TS-LSTM) framework composed of multiple parts, containing short-term, medium-term, and long-term TS-LSTM networks.
IndRNN \cite{li2019deep} not only addresses the gradient vanishing and exploding issues, but it is also faster than the original LSTM. 

\subsubsection{CNN-Based Methods}
CNNs have achieved great success in 2D image analysis due to their superior capability in learning features in the spatial domain. 
However, when facing skeleton-based HAR, the modeling of spatio-temporal information becomes a challenge. 
Nevertheless, plenty of advanced approaches have been proposed, which apply temporal convolution on skeleton data \cite{kim2017interpretable}, or represent skeleton sequences as pseudo-images that are then fed to standard CNNs for HAR \cite{hou2016skeleton, wang2016action}. 
These pseudo-images were designed to simultaneously encode spatial structure information in each frame and temporal dynamic information between frames. 

Hou et al. \cite{hou2016skeleton} and Wang et al. \cite{wang2016action} respectively proposed the skeleton optical spectra and the joint trajectory maps. They both encoded the spatio-temporal information of skeleton sequences into color-texture images, and then adopted CNNs for HAR. 
As an extension of these works, the Joint Distance Map (JDM) \cite{li2017joint} produces view-invariant color-texture images encoding pair-wise distances of skeleton joints.
Ke et al. \cite{ke2017new} transformed each skeleton sequence into three ``video clips'', which were then fed to a pre-trained CNN to produce a compact representation, followed by a multi-task learning network for HAR. 
Kim and Reiter \cite{kim2017interpretable} utilized the Temporal CNN (TCN) \cite{Lea2017TemporalCN} explicitly providing a type of interpretable spatio-temporal representations. 
An end-to-end framework to learn co-occurrence features with a hierarchical methodology was proposed in \cite{li2018co}. 
Specifically, they learned the point-level features of each joint independently, and then utilized these features as a channel of the convolutional layer to learn hierarchical co-occurrence features, and a two-stream framework was adopted to fuse the motion features. 
Caetano et al. introduced SkeleMotion \cite{caetano2019skelemotion} and the Tree Structure Reference Joints Image (TSRJI) \cite{caetano2019skeleton} as representations of skeleton sequences for HAR. 
In \cite{liu2019skepxels}, Skepxels were utilized as basic building blocks to construct skeletal images mainly encoding the spatio-temporal information of human joint locations and velocities. 
Besides, Skepxels were used to hierarchically capture the micro-temporal relations between the joints in the frames, and Fourier Temporal Pyramids \cite{rahmani20163d} were utilized to exploit the macro-temporal relations. 

Plenty of researches focused on addressing certain specific problems. 
For example, the works of \cite{li2017joint, liu2017enhanced, zhang2019view} focused on handling the viewpoint variation issue. 
Since features learned from skeleton data are not always translation, scale, and rotation invariant, several methods \cite{ke2017skeletonnet, li2017skeleton} have also been designed to handle these issues. 
CNN architectures are often complex, resulting in high computational costs. 
Thus Yang et al. \cite{yang2019make} proposed a Double-feature Double-motion Network (DD-Net) to make the CNN-based recognition models run faster. 
Additionally, Tang et al. \cite{tang2022learning} proposed a self-supervised learning framework under the unsupervised domain adaptation setting, which segments and permutes the time segments or body parts to reduce domain-shift and improve the generalization ability of the model. 

\begin{figure}[t]
\centering
\subfigure[RNN-based method.]{
\begin{minipage}[t]{0.4\linewidth}
\centering
\includegraphics[width=1\textwidth]{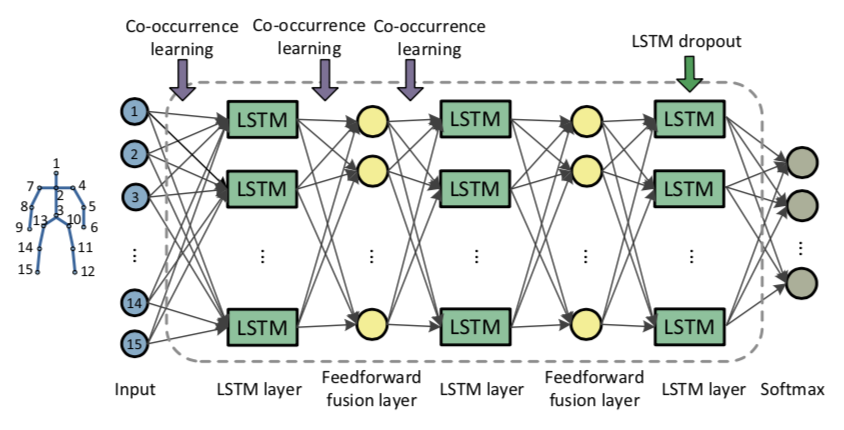}
\label{fig:zhu2016co}
\end{minipage}%
}%
\subfigure[CNN-based method.]{
\begin{minipage}[t]{0.6\linewidth}
\centering
\includegraphics[width=1\textwidth]{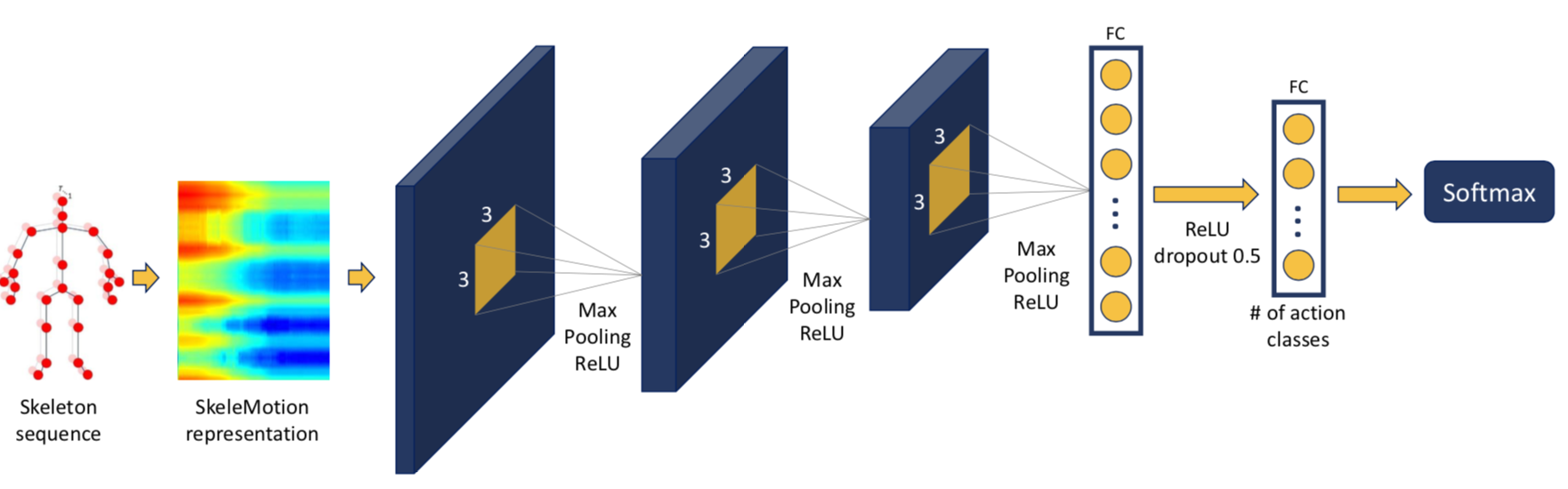}
\label{fig:caetano2019skelemotion}
\end{minipage}%
}%

\subfigure[GCN-based method.]{
\begin{minipage}[t]{1\linewidth}
\centering
\includegraphics[height=0.7in]{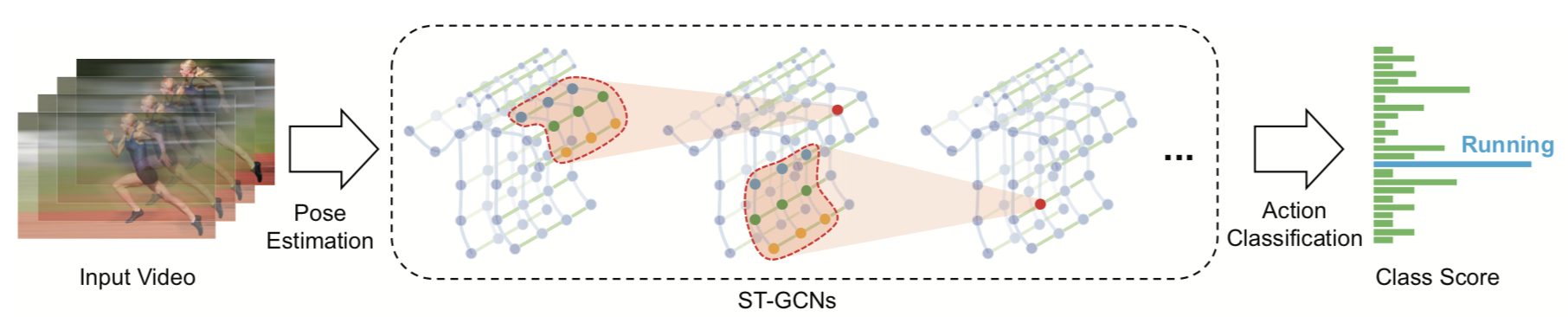}
\label{fig:yan2018spatial}
\end{minipage}%
}%

\subfigure[Transformer-based method.]{
\begin{minipage}[t]{1\linewidth}
\centering
\includegraphics[height=0.65in]{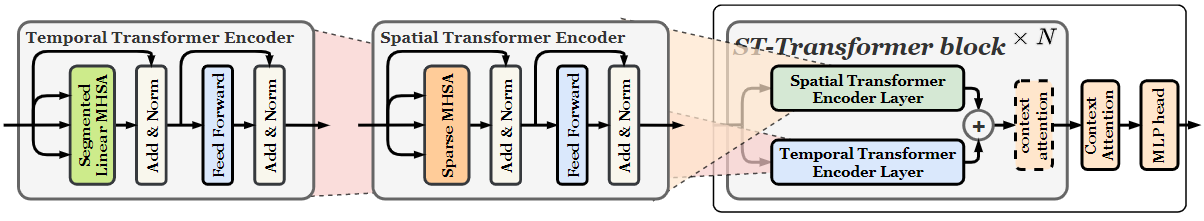}
\label{fig:shi2021star}
\end{minipage}%
}%
\centering
\vspace{-3mm}
\caption{\footnotesize Illustration of deep learning frameworks for skeleton-based HAR. 
(a) Skeleton sequences can be processed by RNN and LSTM as time series. 
(b) Skeleton sequences can be converted to 2D pseudo-images and then be fed to 2D CNNs for feature learning. 
(c) The joint dependency structure can naturally be represented via a graph structure, and thus GCN models are also suitable for this task. 
(d) Transformers model spatio-temporal correlations of skeleton sequences.
(a)-(d) are originally shown in \cite{zhu2016co, caetano2019skelemotion, yan2018spatial, shi2021star}.}
\label{fig:skeleton}
\end{figure}

\subsubsection{GNN or GCN-Based Methods}
Due to the expressive power of graph structures, analyzing graphs with learning models have received great attention recently \cite{zhao2019t, monti2018motifnet}. 
As shown in Fig.~\ref{fig:yan2018spatial}, skeleton data is naturally in the form of graphs. 
Hence, simply representing skeleton data as a vector sequence processed by RNNs, or 2D/3D maps processed by CNNs, cannot fully model the complex spatio-temporal configurations and correlations of the body joints. 
This indicates that topological graph representations can be more suitable for representing the skeleton data. 
As a result, many GNN and GCN-based HAR methods \cite{yan2018spatial, si2018skeleton} have been proposed to treat the skeleton data as graph structures of edges and nodes. 

GNN is a connection model capturing the dependence within a graph through message passing between nodes. 
Si et al. \cite{si2018skeleton} proposed a spatial reasoning network, followed by RNNs, to capture the high-level spatial structural and temporal dynamics of skeleton data. 
Shi et al. \cite{shi2019skeleton} represented the skeleton as a directed acyclic graph to effectively incorporate both the joint and bone information, and utilized a GNN to perform HAR. 

More recently, GCN-based HAR has become a hot research direction \cite{tang2018deep, zhang2019graph, korban2020ddgcn, chengdecoupling, yu2020structure}. 
Yan et al. \cite{yan2018spatial} exploited GCNs for skeleton-based HAR by introducing Spatial-Temporal GCNs (ST-GCNs) that can automatically learn both the spatial and temporal patterns from skeleton data, as shown in Fig. \ref{fig:yan2018spatial}. 
More specifically, the pose information was estimated from the input videos and then passed through the spatio-temporal graphs to achieve action representations with strong generalization capabilities for HAR. Since implicit joint correlations have been ignored by previous works \cite{yan2018spatial}, Li et al. \cite{li2019actional} further proposed an Actional-Structural GCN (AS-GCN), combining actional links and structural links into a generalized skeleton graph. The actional links were used to capture action-specific latent dependencies and the structural links were used to represent higher-order dependencies. 
To better explore the implicit joint correlations, Peng et al. \cite{peng2020learning} determined their GCN architecture via a neural architecture search scheme. Specifically, they enriched the search space to implicitly capture the joint correlations based on multiple dynamic graph sub-structures and higher-order connections with a Chebyshev polynomial approximation. 
Besides, context information integration was used to effectively model long-range dependencies in the work of \cite{zhang2020context}. 

Shi et al. \cite{shi2019two} proposed a two-stream Adaptive GCN (2s-AGCN), where the topology of graphs can be either uniformly or individually learned by the back-propagation algorithm, instead of setting it manually. 
The 2s-AGCN explicitly combined the second-order information (lengths and directions of human bones) of the skeleton with the first-order information (coordinates of joints). 
Wu et al. \cite{wu2019spatial} introduced a cross-domain spatial residual layer to capture the spatio-temporal information, and a dense connection block to learn global information based on ST-GCN. 
Li et al. \cite{li2019spatio} fed frame-wise skeleton and node trajectories of a skeleton sequence to a spatial graph router and a temporal graph router to generate new skeleton-joint-connectivity graphs, followed by a ST-GCN for classification. 
Liu et al. \cite{liu2020disentangling} integrated a disentangled multi-scale aggregation scheme and a spatio-temporal graph convolutional operator named G3D to achieve a powerful feature extractor. 
High-level semantics of joints were introduced for HAR in \cite{zhang2020semantics}. 
Attention mechanisms were applied for extracting discriminative information and global dependencies in \cite{si2019attention, wen2019graph}. 
Besides, to reduce computational costs of GCNs, a Shift-GCN was designed by Cheng et al. \cite{cheng2020skeleton}, which adopts shift graph operations and lightweight point-wise convolutions, instead of using heavy regular graph convolutions. 
\textcolor{black}{Following this research line, Song et al. \cite{song2022constructing} proposed a multi-stream GCN model, which fuses the input branches including joint positions, motion velocities, and bone features at early stage, and utilized separable convolutional layers and a compound scaling strategy to extremely reduce the redundant trainable parameters while increasing the capacity of model.
}
\textcolor{black}{Different from the above mentioned methods, Li et al. \cite{li2021symbiotic} proposed symbiotic GCNs to handle both action recognition and motion prediction tasks simultaneously. The proposed Sym-GNN consists of a multi-branch multi-scale GCN followed by an action recognition and a motion prediction heads, to jointly conduct action recognition and motion prediction tasks. This allows the two tasks to enhance each other. 
}

Recently, Chen et al. \cite{chen2021channel} proposed a Channel-wise Topology Refinement Graph Convolution (CTR-GC) for dynamic topology and multi-channel feature modeling. 
In particular, CTR-GC takes a shared topology matrix as the generic prior for channels, then refines it with inferring channel-specific correlation to acquire channel-wise topologies. Chi et al. \cite{chi2022infogcn} proposed InfoGCN, which comprises an information bottleneck objective to learn maximally informative action representations, and an attention-based graph convolution to infer the context-related skeleton topology. Li et al. \cite{li2021else} proposed an Elastic Semantic Network (Else-Net), which consists of a GCN backbone model followed by multiple layers of elastic units for Continual HAR. Particularly, each elastic unit contains several learning blocks to learn diversified knowledge from different human actions over time, and a switch block to select the most relevant block with respect to the new incoming action.

\subsubsection{Transformer-Based Methods}
As discussed in Section \ref{sec:Transformer}, Transformers have shown great potential in sequential data processing. 
Thus, plenty of methods \cite{zhang2021stst, plizzari2021spatial} applied Transformers on skeleton sequences, with an emphasis on spatio-temporal modeling. 
Zhang et al. \cite{zhang2021stst} proposed a Spatial-Temporal Specialized Transformer (STST), consisting of a spatial Transformer block to model pose information at frame level and a directional temporal Transformer module to capture long dynamics in the temporal dimension. A multi-task self-supervision module was also designed to enhance the robustness of the model to noise. Plizzari et al. \cite{plizzari2021spatial} introduced a Spatial-Temporal Transformer network (ST-TR), where the spatial and temporal self-attention modules learn intra-frame joint interactions and inter-frame motion dynamics, respectively. 
Qiu et al. \cite{qiu2022spatio} presented a Spatio-Temporal Tuples Transformer (STTFormer) architecture for HAR. 
Specifically, STTFormer first divides a skeleton sequence into several non-overlapping clips, and then utilizes the spatio-temporal self-attention module to capture multi-joint dependencies between adjacent frames. Finally, an inter-frame feature aggregation module is employed to aggregate sub-actions. 
Shi et al. \cite{shi2021star} proposed a Sparse Transformer for Action Recognition (STAR), capable of processing varying-length skeleton input without extra pre-processing. 
STAR mainly relies on two modules including a sparse self-attention module to learn spatial relationships via sparse matrix multiplications, and a segmented linear self-attention module to model joint correlations in the temporal dimension.

Cheng et al. \cite{cheng2021hierarchical} presented a hierarchical Transformer for unsupervised skeleton-based HAR, along with a motion prediction pre-training task between adjacent frames to learn discriminative representations. 
Liu et al.\cite{liu2022graph} proposed a Kernel Attention Adaptive Graph Transformer Network (KA-AGTN), which is mainly composed of a skeleton graph transformer block to effectively capture the varying degrees of higher-order dependencies among joints, a temporal kernel attention module, and an adaptive graph strategy. 
To decrease the huge computation and memory cost, Wang et al. \cite{wang2021iip} proposed IIP-Transformer that exploits part-level skeleton data encoding for HAR. 
Specifically, IIP-Transformer regards body joints as five parts, and leverages their relationships to capture intra- and inter-part level dependencies.

In summary, the skeleton modality provides the body structure information, which is simple, efficient, and informative for representing human behaviors. 
Nevertheless, HAR using skeleton data still faces challenges, due to its very sparse representation, the noisy skeleton information, and the lack of shape information that can be important when handling human-object interactions. 
Hence, some of the existing works on HAR also focused on using depth maps, as discussed in the following section, since depth maps not only provide the 3D geometric structural information but also conserve the shape information. 

\begin{table}[t]
\caption{\footnotesize Performance of skeleton-based deep learning HAR methods on NTU RGB+D and NTU RGB+D 120 datasets. `CS', `CV', and `CP' denote Cross-Subject, Cross-View, and Cross-Setup evaluation criteria.}
\vspace{-3mm}
\centering
\resizebox{0.47\textwidth}{!}{
\setlength{\tabcolsep}{7.0pt}
\begin{tabular}{clcccccc}
\toprule[1pt]
\multicolumn{2}{c}{\multirow{3}{*}{\textbf{Methods}}} &
  \multirow{3}{*}{\textbf{~~Year~~}} &
  \multicolumn{4}{c}{\textbf{Dataset}} \\ \cline{4-7} 
\multicolumn{2}{c}{} &
   &
  \multicolumn{2}{c}{\textbf{NTU RGB+D}} &
  \multicolumn{2}{c}{\textbf{NTU RGB+D 120}} \\
\cmidrule(lr){4-5}\cmidrule(lr){6-7}
\multicolumn{2}{c}{}                               &                                      & \textbf{~~~CS~~~} & \textbf{~~CV~~~~} & \textbf{~~CS~~} & \textbf{~~CP~~} \\ \hline
\multicolumn{1}{c|}{\multirow{16}{*}{\rotatebox{90}{\textbf{RNN-Based}}}} & dRNN                   \cite{veeriah2015differential} & 2015 & -           & -           & -           & -           \\ 
\multicolumn{1}{c|}{}                            & HBRNN-L                  \cite{du2015hierarchical}      & 2015 & 59.1        & 64.0        & -           & -           \\ 
\multicolumn{1}{c|}{}                            & Co-occurrence LSTM     \cite{zhu2016co}               & 2016 & -           & -           & -           & -           \\ 
\multicolumn{1}{c|}{}                            & 2 Layer P-LSTM        \cite{shahroudy2016ntu}        & 2016 & 62.9        & 70.3        & 25.5        & 26.3        \\ 
\multicolumn{1}{c|}{} &  Trust Gate ST-LSTM \cite{liu2017skeleton} &  2016 &  69.2 &  77.7 &  58.2 &  60.9 \\ 
\multicolumn{1}{c|}{}                            & Zhang et al.          \cite{zhang2017geometric}      & 2017 & 70.3        & 82.4        & -           & -           \\ 
\multicolumn{1}{c|}{}                            & Two-stream RNN         \cite{wang2017modeling}        & 2017 & 71.3        & 79.5        & -           & -           \\ 
\multicolumn{1}{c|}{}                            & STA-LSTM               \cite{song2017end}             & 2017 & 73.4        & 81.2        & -           & -           \\ 
\multicolumn{1}{c|}{}                            & GCA-LSTM               \cite{liu2017global}           & 2017 & 74.4        & 82.8        & 58.3        & 59.2        \\ 
\multicolumn{1}{c|}{}                            & Ensemble TS-LSTM       \cite{lee2017ensemble}         & 2017 & 74.6        & 81.3        & -           & -           \\ 
\multicolumn{1}{c|}{}                            & VA-LSTM                \cite{zhang2017view}           & 2017 & 79.4        & 87.6        & -           & -           \\ 
\multicolumn{1}{c|}{}                            & Zhang et al.                \cite{zhang2018fusing}           & 2018 & 76.4        & 87.7        & -           & -           \\ 
\multicolumn{1}{c|}{}                            & MANs                      \cite{xie2018memory}           & 2018 & 82.7        & 93.2        & -           & -           \\ 
\multicolumn{1}{c|}{}                           & VA-RNN (aug.)              \cite{zhang2019view}           & 2019 & 79.8        & 88.9        & -           & -           \\ 
\multicolumn{1}{c|}{}                            & dense-IndRNN-aug                 \cite{li2019deep}              & 2019 & 86.7        & 93.7        & -           & -           \\ 
\multicolumn{1}{c|}{}                            & \textcolor{black}{KShapeNet}                 \cite{friji2021geometric}              & 2021 & 97.0        & 98.5        & 90.6           & 86.7           \\ \hline 
\multicolumn{1}{c|}{\multirow{19}{*}{\rotatebox{90}{\textbf{CNN-Based}}}} & Du et al.             \cite{du2015skeleton}          & 2015 & -           & -           & -           & -           \\ 
\multicolumn{1}{c|}{}                            & Hou et al.            \cite{hou2016skeleton}         & 2016 & -           & -           & -           & -           \\ 
\multicolumn{1}{c|}{}                            & JTM                    \cite{wang2016action}          & 2016 & 73.4        & 75.2        & -           & -           \\ 
\multicolumn{1}{c|}{}                            & Res-TCN                    \cite{kim2017interpretable}    & 2017 & 74.3        & 83.1        & -           & -           \\ 
\multicolumn{1}{c|}{}                            & SkeletonNet            \cite{ke2017skeletonnet}       & 2017 & 75.9        & 81.2        & -           & -           \\ 
\multicolumn{1}{c|}{}                            & JDM                    \cite{li2017joint}             & 2017 & 76.2        & 82.3        & -           & -           \\ 
\multicolumn{1}{c|}{}                            & Clips+CNN+MTLN           \cite{ke2017new}               & 2017 & 79.6        & 84.8        & 58.4        & 57.9        \\ 
\multicolumn{1}{c|}{}                            & Liu et al.     \cite{liu2017enhanced}         & 2017 & 80.0        & 87.2        & 60.3        & 63.2        \\ 
\multicolumn{1}{c|}{}                            & Li et al.             \cite{li2017skeleton}          & 2017 & 85.0        & 92.3        & -           & -           \\ 
\multicolumn{1}{c|}{}                            & RotClips+MTCNN         \cite{ke2018learning}          & 2018 & 81.1        & 87.4        & 62.2        & 61.8        \\ 
\multicolumn{1}{c|}{}                            & Xu et al.             \cite{xu2018ensemble}          & 2018 & 84.8        & 91.2        & -           & -           \\ 
\multicolumn{1}{c|}{}                            & HCN                    \cite{li2018co}                & 2018 & 86.5        & 91.1        & -           & -           \\ 
\multicolumn{1}{c|}{}                            & DD-Net                 \cite{yang2019make}            & 2019 & -           & -           & -           & -           \\ 
\multicolumn{1}{c|}{}                            & TSRJI                  \cite{caetano2019skeleton}     & 2019 & 73.3        & 80.3        & 67.9        & 62.8        \\ 
\multicolumn{1}{c|}{}                            & SkeleMotion            \cite{caetano2019skelemotion}  & 2019 & 76.5        & 84.7        & 67.7        & 66.9        \\ 
\multicolumn{1}{c|}{}                            & Skepxel                \cite{liu2019skepxels}         & 2019 & 81.3        & 89.2        & -           & -           \\ 
\multicolumn{1}{c|}{}                            & Li et al.             \cite{li2019learning}          & 2019 & 82.9        & 90.0        & -           & -           \\ 
\multicolumn{1}{c|}{}                            & VA-CNN (aug.)                  \cite{zhang2019view}              & 2019 & 88.7        & 94.3        & -           & -           \\
\multicolumn{1}{c|}{}                             & Banerjee et al.             \cite{banerjee2020fuzzy}           & 2020 & 84.2        & 89.7        & 74.8           & 76.9           \\ \hline 
\multicolumn{1}{c|}{\multirow{26}{*}{\rotatebox{90}{\textbf{GNN or GCN-Based}}}} & ST-GCN                 \cite{yan2018spatial}          & 2018 & 81.5        & 88.3        & -           & -           \\ 
\multicolumn{1}{c|}{}    & DPRL                   \cite{tang2018deep}            & 2018 & 83.5        & 89.8        & -           & -           \\ 
\multicolumn{1}{c|}{}  & SR-TSL                 \cite{si2018skeleton}          & 2018 & 84.8        & 92.4        & -           & -           \\ 
\multicolumn{1}{c|}{}                            & motif-GCNs          \cite{wen2019graph}            & 2019 & 84.2        & 90.2        & -           & -           \\ 
\multicolumn{1}{c|}{}                            & BPLHM                  \cite{zhang2019graph}          & 2019 & 85.4        & 91.1        & -            & -             \\ 
\multicolumn{1}{c|}{}                            & AS-GCN                 \cite{li2019actional}          & 2019 & 86.8        & 94.2        & -           & -           \\ 
\multicolumn{1}{c|}{}                            & STGR-GCN                   \cite{li2019spatio}            & 2019 & 86.9        & 92.3        & -           & -           \\ 
\multicolumn{1}{c|}{}                            & 2s-AGCN                \cite{shi2019two}              & 2019 & 88.5        & 95.1        & -           & -           \\ 
\multicolumn{1}{c|}{}                            & AGC-LSTM  \cite{si2019attention}         & 2019 & 89.2        & 95.0        & -           & -           \\ 
\multicolumn{1}{c|}{}                            & 2s-SDGCN               \cite{wu2019spatial}           & 2019 & 89.6        & 95.7        & -           & -           \\ 
\multicolumn{1}{c|}{}                            & DGNN                   \cite{shi2019skeleton}         & 2019 & 89.9        & 96.1        & -           & -           \\ 
\multicolumn{1}{c|}{}                            & Advanced CA-GCN        \cite{zhang2020context}        & 2020 & 83.5        & 91.4        & -           & -           \\ 
\multicolumn{1}{c|}{}                            & DC-GCN+ADG                    \cite{chengdecoupling}      & 2020 & 88.2        & 95.2        & 90.8        & 96.6        \\ 
\multicolumn{1}{c|}{}                            & SGN                    \cite{zhang2020semantics}      & 2020 & 89.0        & 94.5        & 79.2        & 81.5        \\ 
\multicolumn{1}{c|}{}                            & GCN-NAS                \cite{peng2020learning}        & 2020 & 89.4        & 95.7        & -           & -           \\ 
\multicolumn{1}{c|}{}                            & 4s Shift-GCN           \cite{cheng2020skeleton}       & 2020 & 90.7        & 96.5        & 85.9        & 87.6        \\ 
\multicolumn{1}{c|}{}                            & DDGCN                 \cite{korban2020ddgcn}    & 2020 & 91.1        & 97.1        & -        & - \\
\multicolumn{1}{c|}{}                            & MS-G3D Net                \cite{liu2020disentangling}    & 2020 & 91.5        & 96.2        & 86.9        & 88.4        \\ 
\multicolumn{1}{c|}{}                            & \textcolor{black}{3s-CrosSCLR (FT)}                \cite{li20213d}    & 2021 & 86.2        & 92.5        & 80.5        & 80.4       \\ 
\multicolumn{1}{c|}{}                            & \textcolor{black}{Sym-GNN}                \cite{li2021symbiotic}    & 2021 & 90.1        & 96.4        & -        & -       \\ 
\multicolumn{1}{c|}{}                            & \textcolor{black}{3s-AdaSGN}                \cite{shi2021adasgn}    & 2021 & 90.5        & 95.3        & 85.9        & 86.8       \\ 
\multicolumn{1}{c|}{}                            & \textcolor{black}{Else-Net}                \cite{li2021else}    & 2021 & 91.6        & 96.4        & -        & -       \\ 
\multicolumn{1}{c|}{}                            & \textcolor{black}{CTR-GCN}                \cite{chen2021channel}    & 2021 & 92.4        & 96.8        & 88.9        & 90.6       \\ 
\multicolumn{1}{c|}{}                            & \textcolor{black}{SATD-GCN}               \cite{zhang2022spatial}    & 2022 & 89.3        & 95.5        & -        & -       \\ 
\multicolumn{1}{c|}{}                            & \textcolor{black}{EfficientGCN-B4}               \cite{song2022constructing}    & 2022 & 92.1        & 96.1        & 88.7        & 88.9       \\ 
\multicolumn{1}{c|}{}                            & \textcolor{black}{InfoGCN}               \cite{chi2022infogcn}    & 2022 & 93.0        & 97.1        & 89.8        & 91.2       \\ 
\hline
\multicolumn{1}{c|}{\multirow{7}{*}{\rotatebox{90}{\textbf{Transformer}}}} & \textcolor{black}{Cheng et al.}               \cite{cheng2021hierarchical}    & 2021 & 69.3        & 72.8        & -        & -       \\ 
\multicolumn{1}{c|}{}    &    \textcolor{black}{STAR (sparse)}                 \cite{shi2021star}          & 2021 & 83.4       & 84.2        & 78.3           & 78.5           \\
\multicolumn{1}{c|}{}    &   \textcolor{black}{ST-TR}               \cite{plizzari2021spatial}    & 2021 & 89.9        & 96.1        & 81.9        & 84.1       \\                       
\multicolumn{1}{c|}{}                            &  \textcolor{black}{STST}                 \cite{zhang2021stst}          & 2021 & 91.9       & 96.8        & -           & -           \\
\multicolumn{1}{c|}{}                            &  \textcolor{black}{IIP-Transformer}                 \cite{wang2021iip}          & 2021 & 92.3       & 96.4        & 88.4           & 89.7           \\
\multicolumn{1}{c|}{}                            &  \textcolor{black}{KA-AGTN (2s)}                 \cite{liu2022graph}          & 2022 & 90.4       & 96.1       & 86.1           & 88.0          \\
\multicolumn{1}{c|}{}                            &  \textcolor{black}{STTFormer}                 \cite{qiu2022spatio}          & 2022 & 92.3       & 96.5       & 88.3           & 89.2          \\
                            \bottomrule[1pt]
\end{tabular}%
}
\label{3}
\end{table}

\subsection{DEPTH MODALITY}
Depth maps refer to images where the pixel values represent the distance information from a given viewpoint to the points in the scene. 
The depth modality, which is often robust to variations of color and texture, provides reliable 3D structural and geometric shape information of human subjects, and thus can be used for HAR. The essence of constructing a depth map is to convert the 3D data into a 2D image, and different types of devices have been developed to obtain depth images, which include active sensors (e.g., Time-of-Flight and structured-light-based cameras) and passive sensors (e.g., stereo cameras) \cite{chen2013survey}. Active sensors emit radiation in the direction towards the objects in the scene, and then measure the reflected energy from the objects to acquire the depth information. 
In contrast, passive sensors measure the natural energy that is emitted or reflected by the objects in the scene. 
For example, as a type of passive sensor, stereo cameras usually have two or more lenses to simulate the binocular vision of human eyes to acquire depth information, and depth maps are recovered by seeking image point correspondences between stereo pairs \cite{mohaghegh2018feat}. 
Compared to active sensors (e.g., Kinect and RealSense3D), passive depth map generation is usually computationally expensive, and moreover, depth maps obtained with passive sensors can be ineffective in texture-less regions or highly textured regions with repetitive patterns. 

In this section, we review methods that used depth maps for HAR. Note that only a few works \cite{harville2004fast, roh2010view} used depth maps captured by stereo cameras for HAR, while most of the other methods \cite{wang2017structured, rahmani20163d, wang2018depth} focused on using depth videos captured by active sensors to recognize actions, due to the availability of low-cost and reliable active sensors like Kinect. Consequently, here we focus on reviewing the methods using depth videos captured with active sensors for HAR. 
Table \ref{4} shows the results of some depth-based HAR methods on several benchmark datasets. 

Compared to hand-crafted feature-based methods \cite{yang2014super, oreifej2013hon4d}, deep learning models \cite{rahmani20163d, wang2017structured} have shown to be more powerful and achieved better performance for HAR from depth maps. 
Inspired by the hand-crafted Depth Motion Map (DMM) features \cite{yang2012recognizing}, a deep learning framework utilizing weighted hierarchical DMMs was proposed in \cite{wang2015action2}.
As DMMs are not able to capture detailed temporal information, Wang et al. \cite{wang2017structured} proposed to represent depth sequences with three pairs of structured dynamic images \cite{fernando2016rank} at the body, body part, and joint levels, which were then fed to CNNs followed by a score fusion module for fine-grained HAR. 

The performance of this method was further improved in \cite{wang2018depth} by introducing three representations of depth maps, including dynamic depth images, dynamic depth normal images, and dynamic depth motion normal images. 
Rahmani et al. \cite{rahmani20163d} transferred the human data obtained from different views to a view-invariant high-level space to address the view-invariant HAR problem. 
Specifically, they utilized a CNN model to learn the view-invariant human pose model and Fourier Temporal Pyramids to model the temporal action variations. 
To obtain more multi-view training data, synthetic data was generated by fitting synthetic 3D human models to real motion capture data, and rendering the human data from various viewpoints. In \cite{xiao2019action}, multi-view dynamic images were extracted through multi-view projections from depth videos for action recognition. 
In order to effectively capture spatio-temporal information in depth videos, Sanchez et al. \cite{sanchez20203dfcnn} proposed a 3D fully CNN architecture for HAR. In their subsequent work \cite{sanchez2020exploiting}, a variant of LSTM unit was introduced to address the problem of memory limitation during video processing, which can be used to perform HAR from long and complex videos. 

Note that besides using depth maps obtained with active sensors or stereo cameras for HAR, there has been another approach designed for depth-based HAR from RGB videos. Specifically, Zhu and Newsam \cite{zhu2016depth2action} estimated the depth maps from RGB videos using existing depth estimation techniques \cite{LiuSL15deep, Eigen15deep}, which were then passed through a deep learning architecture for action classification. 

In general, the depth modality provides geometric shape information that is useful for HAR. 
However, the depth data is often not used alone, due to the lack of appearance information that can also be helpful for HAR in some scenarios. 
Thus, many works focused on fusing depth information with other data modalities for enhanced HAR, and more details can be found in Section \ref{MULTI-MODALITY}. 

\subsection{INFRARED MODALITY}
Generally, infrared sensors do not need to rely on external ambient light, and thus are particularly suitable for HAR at night. 
Infrared sensing technologies can be divided into active and passive ones. 
Some infrared sensors, such as Kinect, rely on active infrared technology, which emit infrared rays and utilize target reflection rays to perceive objects in the scene. 
In contrast, thermal sensors relying on passive infrared technology do not emit infrared rays. Instead, they work by detecting rays (i.e., heat energy) emitted from targets. 

Some deep learning methods \cite{gao2016infar, akula2018deep} have been proposed for HAR from infrared data. 
In \cite{kawashima2017action}, the extremely low-resolution thermal images were first cropped by the gravity center of human regions. The cropped sequences and the frame differences were then passed through a CNN followed by an LSTM layer to model spatio-temporal information for HAR. 
To simultaneously learn both the spatial and temporal features from thermal videos, Shah et al. \cite{shah2018spatio} utilized a 3D CNN and achieved real-time HAR. 
Instead of using raw thermal images, Meglouli et al. \cite{meglouli2019new} passed the optical flow information computed from thermal sequences into a 3D CNN for HAR. 

Inspired by the two-stream CNN model \cite{simonyan2014two}, several multi-stream architectures \cite{gao2016infar, jiang2017learning, liu2018global} have also been proposed. 
Gao et al. \cite{gao2016infar} passed optical flow and optical flow motion history images (OF-MHIs) \cite{TsaiCL15} through a two-stream CNN for HAR. 
In \cite{jiang2017learning}, both infrared data and optical flow volumes were fed to a two-stream 3D CNN framework to effectively capture the complementary information on appearance from still thermal frames and motion between frames. The two-stream model was trained using a combination of cross-entropy and label consistency regularization loss in \cite{Jiang2017LearningDF}. 
To better represent the global temporal information, Optical flow motion history images (OF-MHIs) \cite{TsaiCL15}, optical flow and a stacked difference of optical flow images, were fed to a three-stream CNN for HAR \cite{liu2018global}.
Imran and Raman \cite{imran2019deep} proposed a four-stream architecture, where each stream consists of a CNN followed by an LSTM. The local/global stacked dense flow difference images \cite{imran2020evaluating} and local/global Stacked saliency difference images were fed to these four streams to capture both the local and global spatio-temporal information in videos. 
Mehta et al. \cite{mehta2020motion} presented an adversarial framework consisting of a two-stream 3D convolutional auto-encoder as the generator and two 3D CNNs as the joint discriminator. The generator network took thermal data and optical flow as inputs, and the joint discriminator tried to discriminate the real thermal data and optical flow from the reconstructed ones. 

In all, infrared data has been used for HAR, especially for night HAR. However, infrared images may suffer from a relatively low contrast and a low signal-to-noise ratio, making it challenging for robust HAR in some scenarios. 

\begin{table}
\caption{Performance of depth, infrared, point cloud, and event stream-based deep learning HAR methods. The datasets are MSRDailyActivity3D (M) \cite{wang2012mining}, Northwestern-UCLA (N-UCLA) \cite{wang2014cross}, UWA3D Multiview II (UWA3D II) \cite{rahmani2016histogram}, NTU RGB+D (N)\cite{shahroudy2016ntu}, InfAR \cite{gao2016infar}, MSR-Action3D \cite{li2010action}, DvsGesture \cite{amir2017low}, and DHP19 \cite{calabrese2019dhp19}.}
\vspace{-3mm}
\centering
\setlength{\tabcolsep}{3.5pt}
\resizebox{0.47\textwidth}{!}{%
\begin{tabular}{clcccccc}
\toprule[1pt]
\multicolumn{2}{c}{\textbf{Method}} &
  \textbf{Year} &
  \multicolumn{5}{c}{\textbf{Dataset}} \\ \hline
\multicolumn{1}{c|}{\multirow{9}{*}{\rotatebox{90}{\textbf{Depth}}}} &
  \multirow{2}{*}{\textbf{}} &
  \multirow{2}{*}{\textbf{}} &
  \multirow{2}{*}{\textbf{M}} &
  \multirow{2}{*}{\textbf{N–UCLA}} &
  \multirow{2}{*}{\textbf{UWA3D II}} &
  \multicolumn{2}{c}{\textbf{N}} \\ \cline{7-8} 
\multicolumn{1}{c|}{} &
   &
   &
   &
   &
   &
  \textbf{CS} &
  \textbf{CV} \\
\multicolumn{1}{c|}{} &
  Wang et al. \cite{wang2015action2} &
  2015 &
  85.0 &
  - &
  - &
  - &
  - \\
\multicolumn{1}{c|}{} &
  Rahmani et al. \cite{rahmani20163d} &
  2016 &
  80.0 &
  92.0 &
  76.9 &
  - &
  - \\
\multicolumn{1}{c|}{} &
  S$^2$DDI \cite{wang2017structured} &
  2017 &
  100 &
  - &
  - &
  - &
  - \\
\multicolumn{1}{c|}{} &
  Wang et al. \cite{wang2018depth} &
  2018 &
  - &
  - &
  - &
  87.1 &
  84.2 \\
\multicolumn{1}{c|}{} &
  MVDI \cite{xiao2019action} &
  2019 &
  - &
  84.2 &
  68.1 &
  84.6 &
  87.3 \\
\multicolumn{1}{c|}{} &
  3DFCNN \cite{sanchez20203dfcnn} &
  2020 &
  - &
  83.6 &
  66.6 &
  78.1 &
  80.4 \\
\multicolumn{1}{c|}{} &
  Stateful ConvLSTM \cite{sanchez2020exploiting} &
  2020 &
  - &
  - &
  - &
  80.4 &
  79.9 \\ \hline
\multicolumn{1}{c|}{\multirow{10}{*}{\rotatebox{90}{\textbf{Infrared}}}} &
   &
   &
  \multicolumn{5}{c}{\textbf{InfAR}} \\
\multicolumn{1}{c|}{} &
  Gao et al.\cite{gao2016infar} &
  2016 &
  \multicolumn{5}{c}{76.7} \\
\multicolumn{1}{c|}{} &
  Jiang et al. \cite{jiang2017learning} &
  2017 &
  \multicolumn{5}{c}{77.5} \\
\multicolumn{1}{c|}{} &
  Kawashima et al. \cite{kawashima2017action} &
  2017 &
  \multicolumn{5}{c}{-} \\
\multicolumn{1}{c|}{} &
  Shah et al. \cite{shah2018spatio} &
  2018 &
  \multicolumn{5}{c}{-} \\
\multicolumn{1}{c|}{} &
  TSTDDs \cite{liu2018global} &
  2018 &
  \multicolumn{5}{c}{79.3} \\
\multicolumn{1}{c|}{} &
  Akula et al. \cite{akula2018deep} &
  2018 &
  \multicolumn{5}{c}{-} \\
\multicolumn{1}{c|}{} &
  Imran et al.\cite{imran2019deep} &
  2019 &
  \multicolumn{5}{c}{83.5} \\
\multicolumn{1}{c|}{} &
  Meglouli et al.\cite{meglouli2019new} &
  2019 &
  \multicolumn{5}{c}{88.2} \\
\multicolumn{1}{c|}{} &
  Mehta et al. \cite{mehta2020motion} &
  2020 &
  \multicolumn{5}{c}{-} \\ \hline
\multicolumn{1}{c|}{\multirow{12}{*}{\rotatebox{90}{\textbf{Point Cloud}}}} &
  \multirow{2}{*}{} &
  \multirow{2}{*}{} &
  \multicolumn{2}{c}{\multirow{2}{*}{\textbf{MSR-Action3D}}} &
   &
  \multicolumn{2}{c}{\textbf{N}} \\ \cline{7-8} 
\multicolumn{1}{c|}{} &
   &
   &
  \multicolumn{2}{c}{} &
   &
  \textbf{CS} &
  \textbf{CV} \\
\multicolumn{1}{c|}{} &
  \textcolor{black}{PAT} \cite{yang2019modeling} &
  2019 &
  \multicolumn{2}{c}{-} &
   &
  - &
  - \\
  \multicolumn{1}{c|}{} &
  MeteorNet \cite{liu2019meteornet} &
  2019 &
  \multicolumn{2}{c}{88.5} &
   &
  - &
  - \\
\multicolumn{1}{c|}{} &
  PointLSTM \cite{Min_2020_CVPR} &
  2020 &
  \multicolumn{2}{c}{-} &
   &
  - &
  - \\
\multicolumn{1}{c|}{} &
  3DV-PointNet++ \cite{wang20203dv} &
  2020 &
  \multicolumn{2}{c}{-} &
   &
  88.8 &
  96.3 \\
\multicolumn{1}{c|}{} &
  ASTACNN \cite{wang2020anchorbased} &
  2020 &
  \multicolumn{2}{c}{93.0} &
   &
  - &
  - \\ 
\multicolumn{1}{c|}{} &
  4D MinkNet \cite{Wang_2021_WACV} &
  2021 &
  \multicolumn{2}{c}{86.3} &
   &
  - &
  - \\
\multicolumn{1}{c|}{} &
  \textcolor{black}{P4Transformer} \cite{fan2021point} &
  2021 &
  \multicolumn{2}{c}{90.9} &
   &
  90.2 &
  96.4\\ 
  \multicolumn{1}{c|}{} &
  \textcolor{black}{PSTNet} \cite{fan2021pstnet} &
  2021 &
  \multicolumn{2}{c}{91.2} &
   &
  90.5 &
  96.5\\ 
    \multicolumn{1}{c|}{} &
  \textcolor{black}{PST$^2$ (16 frames)} \cite{wei2022spatial} &
  2022 &
  \multicolumn{2}{c}{89.2} &
   &
  - &
  -\\ 
      \multicolumn{1}{c|}{} &
  \textcolor{black}{PST-Transformer} \cite{fan2022point} &
  2022 &
  \multicolumn{2}{c}{93.7} &
   &
  91.0 &
  96.4\\ \hline
\multicolumn{1}{c|}{\multirow{8}{*}{\rotatebox{90}{\textbf{Event Stream}}}} &
   &
   &
  \multicolumn{2}{c}{\textbf{DvsGesture}} &
  \textbf{} &
  \multicolumn{2}{c}{\textbf{DHP19}} \\
\multicolumn{1}{c|}{} &
  Amir et al. \cite{amir2017low} &
  2017 &
  \multicolumn{2}{c}{96.5} &
   &
  \multicolumn{2}{c}{-} \\
\multicolumn{1}{c|}{} &
  Ghosh et al. \cite{ghosh2019spatiotemporal} &
  2019 &
  \multicolumn{2}{c}{94.9} &
   &
  \multicolumn{2}{c}{-} \\
\multicolumn{1}{c|}{} &
  Wang et al. \cite{wang2019space} &
  2019 &
  \multicolumn{2}{c}{97.1} &
   &
  \multicolumn{2}{c}{-} \\
\multicolumn{1}{c|}{} &
  Huang et al. \cite{huang2020event} &
  2020 &
  \multicolumn{2}{c}{97.4} &
   &
  \multicolumn{2}{c}{-} \\
\multicolumn{1}{c|}{} &
  Chen et al. \cite{chen2020dynamic} &
  2020 &
  \multicolumn{2}{c}{98.6} &
   &
  \multicolumn{2}{c}{95.9} \\
\multicolumn{1}{c|}{} &
  Innocenti et al. \cite{innocenti2021temporal} &
  2021 &
  \multicolumn{2}{c}{99.6} &
   &
  \multicolumn{2}{c}{-} \\
\multicolumn{1}{c|}{} &
  \textcolor{black}{E$^2$(GO) \& E$^2$(GO)MO  \cite{plizzari2022e2}} &
  2022 &
  \multicolumn{2}{c}{-} &
   &
  \multicolumn{2}{c}{-} \\
\bottomrule[1pt]
\end{tabular}%
}
\label{4}
\end{table}

\subsection{POINT CLOUD MODALITY}
Point cloud data is composed of a numerous collection of points that represent the spatial distribution and surface characteristics of the target under a spatial reference system. 
There are two main ways to obtain 3D point cloud data, namely, (1) using 3D sensors, such as LiDAR and Kinect, or (2) using image-based 3D reconstruction. 
As a 3D data modality, point cloud has great power to represent the spatial silhouettes and 3D geometric shapes of the subjects, and hence can be used for HAR. 

Early methods \cite{rahmani2014hopc, rahmani2016histogram} focused on extracting hand-crafted spatio-temporal descriptors from the point cloud sequences for HAR, while the current mainstream research focuses on deep learning architectures \cite{fan2021pstnet, fan2021point} that generally achieve better performance. 
Wang et al. \cite{wang20203dv} transformed the raw point cloud sequence into regular voxel sets. Then temporal rank pooling \cite{fernando2016rank} was applied on all the voxel sets to encode 3D action information into one single voxel set. Finally, the voxel representation was abstracted and passed through the PointNet++ model \cite{qi2017pointnet++} for 3D HAR. 
However, converting point clouds into voxel representations causes quantization errors and inefficient processing performance. 
Instead of quantizing the point clouds into voxels, Liu et al. \cite{liu2019meteornet} proposed MeteorNet, which directly stacks
multi-frame point clouds and calculates local features by aggregating information from spatio-temporal neighboring points.
\textcolor{black}{
Unlike MeteorNet \cite{liu2019meteornet}, which 
learns  spatio-temporal information by appending 1D temporal dimension to 3D points, PSTNet \cite{fan2021pstnet} disentangles the space and time to reduce the impacts of the spatial irregularity of points on temporal modeling.
}
In the work of \cite{Wang_2021_WACV}, a self-supervised learning framework was introduced to learn 4D spatio-temporal information from the point cloud sequence by predicting the temporal orders of 4D clips within the sequence. 
A 4D CNN model \cite{liu2019meteornet, choy20194d} followed by an LSTM was utilized to predict the temporal order. Finally, the 4D CNN+LSTM network was fine-tuned on the existing HAR dataset to evaluate its performance. 
Wang et al. \cite{wang2020anchorbased} proposed an anchor-based spatio-temporal attention convolution model to capture the dynamics of 3D point cloud sequences. 
However, these methods are not able to sufficiently capture the long-term relationships within point cloud sequences. 
To address this issue, Min et al. \cite{Min_2020_CVPR} introduced a modified version of LSTM unit named PointLSTM to update state information for neighbor point pairs to perform HAR. 
%

Transformers have also recently gained increasing attention in point cloud-based HAR \cite{yang2019modeling, fan2021point, wei2022spatial, fan2022point}. 
Yang et al. \cite{yang2019modeling} proposed a Point Attention Transformer (PAT), which mainly relies on a lightweight Group Shuffle Attention (GSA) module to learn relationships among points and a permutation-invariant, task-agnostic and differentiable sampling method, named Gumbel Subset Sampling (GSS), to sample a representative subset of input points.
Fan et al. \cite{fan2021point} introduced a point 4D convolution, followed by a Transformer to capture the global appearance and motion information across the entire point cloud video for HAR.
Wei et al. \cite{wei2022spatial} proposed a Point Spatial-Temporal Transformer (PST$^2$) for processing point cloud sequences. 
A self-attention-based module, called Spatio-Temporal Self-Attention (STSA), was introduced to capture the spatial-temporal context information, which can be leveraged for action recognition in 3D point clouds.

In all, point clouds can effectively capture the 3D shapes and silhouettes of subjects, and can be used for HAR. Generally, 3D point cloud-based HAR methods can be insensitive to viewpoint variations, since viewpoint normalization can be conveniently performed by rotating the point cloud in the 3D space. However, point clouds often suffer from the presence of noise and high non-uniformity distribution of points, making it challenging for robust HAR. In addition, processing all the points within the point cloud sequence is often computationally expensive.

\subsection{EVENT STREAM MODALITY}
Event cameras, also known as neuromorphic cameras or dynamic-vision sensors, which can capture illumination changes and produce asynchronous events independently for each pixel \cite{innocenti2021temporal}, have received lots of attention recently. Different from conventional video cameras capturing the entire image arrays, event cameras only respond to changes in the visual scene. Taking a high-speed object as an example, traditional RGB cameras may not be able to capture enough information of the object, due to the low frame rate and motion blur. However, this issue can be significantly mitigated when using event cameras that operate at extremely high frequencies, generating events at a \textmu s temporal scale \cite{innocenti2021temporal}. 
Besides, event cameras have some other characteristics, such as high dynamic range, low latency, low power consumption, and no motion blur, which make them suitable for HAR. Particularly, event cameras are able to effectively filter out background information and keep foreground movement only, avoiding considerable redundancies in the visual information. 
However, the information obtained with event cameras is generally spatio-temporally sparse, and asynchronous. 
Common event cameras include the Dynamic Vision Sensor (DVS) \cite{lichtsteiner2008128} and the Dynamic and Active-pixel Vision Sensor (DAVIS) \cite{berner2013240}. 

The output data of event cameras is highly different from that of conventional RGB cameras, as shown in Table \ref{1}. Thus, some of the existing methods \cite{ghosh2019spatiotemporal, innocenti2021temporal} mainly focused on designing event aggregation strategies converting the asynchronous output of the event camera into synchronous visual frames, which can then be processed with conventional computer vision techniques. While hand-crafted feature-based methods \cite{baby2017dynamic, clady2017motion} have been proposed, deep learning methods have been more popularly used recently. 
Given a fixed time interval $\Delta t$, Innocenti et al. \cite{innocenti2021temporal} first built a sequence of binary representations from the raw event data by checking the presence or absence of an event for each pixel during $\Delta t$. These intermediate binary representations were then stacked together to form a single frame via a binary to decimal conversion. The sequence of such frames extracted from the whole event stream was finally fed to a CNN+LSTM for HAR. 
Huang et al. \cite{huang2020event} utilized time-stamp image encoding to transform the event data sequence into frame-based representations, which were then fed to a CNN for HAR. More precisely, the value of each pixel in the generated time-stamp image encodes the amount of event taking place within a time-window. 
However, since the size of the frame to be processed is basically larger than the original Neuromorphic Vision Stream (NVS), the advantages of event cameras are diluted. 

Therefore, different from the aforementioned methods, some other deep learning approaches \cite{ghosh2019spatiotemporal, george2020reservoir, wang2019space, bi2020graph, chen2020dynamic} have been proposed, which directly take the event data as input of deep neural networks. 
Ghosh et al. \cite{ghosh2019spatiotemporal} learned a set of 3D spatio-temporal convolutional filters in an unsupervised manner to generate a structured matrix form of the raw event data, which was then fed to a 3D CNN for HAR. 
George et al. \cite{george2020reservoir} utilized Spiking Neural Networks (SNNs) for event stream-based HAR. 
The work in \cite{wang2019space} treated the event stream as a 3D point cloud, which was then fed to a PointNet \cite{qi2017pointnet++} for gesture recognition. 
Recently, Bi et al. \cite{bi2020graph} represented events as graphs and used a GCN network for an end-to-end feature learning directly from the raw event data. 
Besides, Plizzari et al. \cite{plizzari2022e2} proposed E$^2$(GO) leveraging traditional CNNs by employing Squeeze And Excitation modules~\cite{hu2018SE} to exploit micro-movements in event data. The E$^2$(GO)MO has also been introduced to distill motion information from optical flow to event data. 

In general, the event stream modality is an emerging modality for HAR, and has received great attention in the past few years. Processing event data is computationally cheap, and the captured frames do not usually contain background information, which can be helpful for action understanding. However, the event stream data cannot generally be effectively and directly processable using conventional video analysis techniques, and thus effectively and directly utilizing the asynchronous event streams for HAR is still a challenging research problem.

\subsection{AUDIO MODALITY}
Audio signal is often available with videos for HAR. Due to the synchronization between the visual and audio streams, the audio data can be used to locate actions to reduce human labeling efforts and decrease computational costs. Some deep learning-based methods \cite{laput2018ubicoustics, liang2019audio} have been proposed to perform general activity recognition from audio signals. However, in recent years, only a few deep learning methods were proposed for HAR from audio signal alone. 
For example, Liang and Thomaz \cite{liang2019audio} used a pre-trained VGGish model \cite{hershey2017cnn} as the feature extractor followed by a deep classification network to perform HAR. 

Using audio modality alone is not a very popular scheme for HAR, compared to other data modalities, since audio signal does not contain enough information for accurate HAR. Nevertheless, the audio modality can serve as complementary information for more reliable and efficient HAR, and thus most of the audio-based HAR methods \cite{owens2018audio, kazakos2019epic, gao2020listen, xiao2020audiovisual} focused on taking advantage of multi-modal deep learning techniques, which are discussed in Section \ref{MULTI-MODALITY}.

\subsection{ACCELERATION MODALITY}
Acceleration signals obtained from accelerometers have been used for HAR \cite{casale2011human}, due to their robustness against occlusion, viewpoint, lighting, and background variations, etc. 
Specifically, a tri-axial accelerometer can return an estimation of acceleration along the $x$, $y$, and $z$ axes, that can be used to perform human activity analysis \cite{bayat2014study}. 
As for the feasibility of using acceleration signal for HAR, although the size and proportion of the human body vary from person to person, people generally have similar qualitative ways to perform an action, so the acceleration signal usually does not have obvious intra-class variations for the same action. 
HAR using acceleration signal can generally achieve a high accuracy, and thus has been adopted for remote monitoring systems \cite{khan2010human, maurer2006activity} while taking care of privacy issues. 

Recently, deep learning networks have been widely used for acceleration-based HAR. 
In \cite{zeng2014convolutional, chen2015deep, panwar2017cnn, ignatov2018real}, the tri-axial acceleration data was fed to different CNN architectures for HAR. 
Wang et al. \cite{wang2019human2} proposed a framework consisting of CNN and Bi-LSTM networks to extract spatial and temporal features from the raw acceleration data. 
Unlike the above-mentioned works, Lu et al. \cite{lu2019robust} utilized a modified Recurrence Plot (RP) \cite{eckmann1995recurrence} to transform the raw tri-axial acceleration data into color images, which were then fed to a ResNet for HAR. 
Besides, some acceleration-based methods \cite{santos2019accelerometer} focused on the fall detection task. 

In general, the acceleration modality can be utilized for fine-grained HAR, and has been used for action monitoring, especially for elderly care due to its privacy-protecting characteristic. However, the subject needs to carry wearable sensors that are often cumbersome and disturbing. In addition, the position of the sensors on the human body can also affect the HAR performance. 

\subsection{RADAR MODALITY}
Radar is an active sensing technology that transmits electromagnetic waves and receives returned waves from targets. 
Continuous-wave radars, such as Doppler \cite{lin2007doppler} and Frequency Modulated Continuous Wave (FMCW) radars \cite{stove2004modern}, are most often chosen for HAR. 
Specifically, Doppler radars detect the radial velocity of the body parts, and the frequency changes according to the distance, which is known as the Doppler shift. 
The micro-Doppler signatures generated by the micro-motion of the radar contain the motion and structure information of the target, which thus can be used for HAR. 
As for the FMCW radars, they can measure the distances of the targets as well. 
There are some advantages of using spectrograms obtained from radars for HAR, which include the robustness to variations in illumination and weather conditions, privacy protection, and the capability of through-wall HAR \cite{li2019survey}.

Several deep learning architectures have been proposed recently. The works in \cite{kim2015human, park2016micro, trommel2016multi, kim2017classification, du2018transfer, shah2019human} fed the micro-Doppler spectrogram images into CNNs for predicting action classes in different scenarios, such as aquatic activities \cite{park2016micro, kim2017classification}, different geometrical locations \cite{shah2019human}, and simulated environments \cite{trommel2016multi}. 
Unlike the aforementioned methods, the work in \cite{lin2018human} directly took the raw radar range data as input. The raw data was passed through an auto-correlation function, followed by a CNN to extract action-related features. Finally, a Random Forest classifier was used to predict the action class. 

The two-stream architecture has also been investigated. 
Hernang{\'o}mez et al. \cite{hernangomez2019human} designed a two-stream CNN taking both micro-Doppler and range spectrograms representing the structural signatures of the target as inputs. 
In the work of \cite{zhou2020two}, micro-Doppler spectrograms and echos of the radar were fed to a two-stream CNN model for HAR. 

By interpreting micro-Doppler spectrograms as temporal sequences rather than images, several RNN-based architectures \cite{wang2019human, yang2019action} have also been proposed recently.
For example, Yang et al. \cite{yang2019action} and Wang et al. \cite{wang2019human} utilized an LSTM model and a stacked RNN model, respectively to predict the action classes. 

In general, the characteristics and advantages of the radar modality make it suitable to be used for HAR in some scenarios, but radars are relatively expensive. Though HAR using radar data has achieved satisfactory results on some datasets, there is still plenty of room for the development of radar-based methods, and the work of \cite{li2019survey} also pointed out some future directions in this area, such as handling more complex actions in real-world scenarios with radar data. 

\subsection{WIFI MODALITY}
WiFi is considered to be one of the most common indoor wireless signal types nowadays \cite{zou2019wifi}. 
Since human bodies are good reflectors of wireless signal, WiFi signal can be utilized for HAR, and sometimes even for through-wall HAR \cite{wu2018tw}. 
There are some advantages of using the WiFi modality for action analysis, mainly due to the convenience, simplicity, and privacy protection of WiFi signal, and also the low cost of WiFi devices. 
Specifically, most of the existing WiFi-based HAR methods \cite{wang2014eyes, wang2017device} focused on using the Channel State Information (CSI) to conduct the HAR task. 
CSI is the fine-grained information computed from the raw WiFi signal, and the WiFi signal reflected by a person who performs an action, usually generates unique variations in the CSI on the WiFi receiver. 

Deep learning-based HAR from the CSI signal has received attention recently. 
Wang et al. \cite{Wang2017wifi} proposed a deep sparse auto-encoder to learn discriminative features from the CSI streams. 
In the work of \cite{wang2019temporal}, the WiFi-based sample-level HAR model, named Temporal Unet, was proposed, which consists of several temporal convolutional, deconvolutional, and max pooling layers.
This method classified every WiFi distortion sample in the series into one action for fine-grained HAR. 
LSTM networks have also been used for HAR using CSI signal \cite{huang2019wiga, sheng2020deep, chen2019mobcom}. 
Huang et al. \cite{huang2019wiga} passed raw CSI measurements through a noise and redundancy removal module followed by a clip reconstruction module to segment the processed CSI signal into multiple clips. These clips were then fed to a multi-stream CNN+LSTM model to perform HAR. 
In the work of \cite{sheng2020deep}, the spatial features of the CSI signal were first extracted from a fully connected layer of a pre-trained CNN. These features were then fed to a Bi-LSTM to capture the temporal information for HAR. 
Chen et al. \cite{chen2019mobcom} directly passed the raw CSI signal through an attention-based Bi-LSTM to predict the action class.
Different from the above-mentioned works, Gao et al. \cite{Gao2017wifi} transformed the CSI signal into radio images, which were then fed to a deep sparse auto-encoder to learn discriminative features for HAR. 


In general, because of the advantages (e.g., convenience), the WiFi modality can be used for HAR in some scenarios. 
However, there are still some challenges which need to be further addressed, such as how to more effectively use the CSI phase and amplitude information, and to improve the robustness when handling dynamic environments. 

Apart from the various modalities mentioned above, some other modalities, such as radio frequency \cite{yang2011activity, li2019making, sun2022recent}, energy harvester \cite{xu2017keh, ma2020simultaneous}, gyroscope \cite{chen2015utd, kong2019mmact}, electromyography \cite{wang2020ev}, and pressure \cite{kong2019mmact} have also been used for HAR.


\section{MULTI-MODALITY} \label{MULTI-MODALITY}
In real life, humans often perceive the environment in a multi-modal cognitive way. 
Similarly, multi-modal machine learning is a modeling approach aiming to process and relate the sensory information from multiple modalities \cite{baltruvsaitis2018multimodal}. 
By aggregating the advantages and capabilities of various data modalities, multi-modal machine learning can often provide more robust and accurate HAR. There are two main types of multi-modality learning methods, namely, fusion and co-learning. 
Fusion refers to the integration of information from two or more modalities for training and inference, while co-learning refers to the transfer of knowledge between different data modalities. 

\begin{table}[t]
\caption{\footnotesize Performance comparison of multi-modality fusion-based HAR methods on MSRDailyActivity3D (M), UTD-MHAD (U), and NTU RGB+D (N) datasets. S: Skeleton, D: Depth, IR: Infrared, PC: Point Cloud, Ac: Acceleration, Gyr: Gyroscope.}
\vspace{-3mm}
\centering
\resizebox{0.4782\textwidth}{!}{
\setlength{\tabcolsep}{0.5pt}
\begin{tabular}{lccccccccc}
\toprule[1pt]
\multirow{3}{*}{\textbf{Method}} &
  \multirow{3}{*}{\textbf{Year}} &
  \multirow{3}{*}{\textbf{~~Modality~~}} &
  \multirow{3}{*}{\textbf{Fusion}} &
  \multicolumn{4}{c}{\textbf{Dataset}} \\ \cline{5-8} 
   &
   &
   &
   &
  \multirow{2}{*}{\textbf{~~M~~}} &
  \multirow{2}{*}{\textbf{~~U~~}} &
  \multicolumn{2}{c}{\textbf{N}} \\\cline{7-8} 
                                                            &                          &             &            &            &          & \textbf{~~CS~~} & \textbf{~~CV~~} \\ \hline
Imran et al.        \cite{imran2016human}          & 2016 &          \multirow{9}{*}{RGB,D}                         & Score                 & -    & 91.2 & -           & -           \\
SFAM                \cite{wang2017scene}           & 2017 &                                   & Feature,Score                 & -    & -    & -           & -           \\
c-ConvNet           \cite{wang2018cooperative}     & 2018 &                                   & Feature,Score                 & -    & -    & 86.4        & 89.1        \\
GMVAR              \cite{wang2019generative}      & 2019 &                                   & Score                & -    & -    & -           & -           \\
Dhiman et al.       \cite{dhiman2020view}          & 2020 &                              & Score                 & -    & -    & 79.4        & 84.1        \\
Wang et al.         \cite{wang2020hybrid}          & 2020 &                              & Feature                 & -    & -    & 89.5        & 91.7        \\
\textcolor{black}{Trear}         \cite{li2021trear}          & 2021 &                              & Feature                 & -    & -    & -        & -        \\
\textcolor{black}{Ren et al.}         \cite{ren2021multi}          & 2021 &                              & Score                 & -    & -    & 89.7        & 93.0        \\
\textcolor{black}{CAPF}         \cite{zhou2022decoupling}          & 2022 &                              & Feature,Score                 & -    & -    & 94.2        & 97.3        \\
\hline
ST-LSTM             \cite{liu2017skeleton}         & 2017 &  \multirow{13}{*}{RGB,S}                                & Feature               & -    & -    & 73.2        & 80.6        \\
Chain-MS            \cite{zolfaghari2017chained}   & 2017 &                                   & Feature                 & -    & -    & 80.8        & -           \\
GRU+STA-Hands      \cite{baradel2017human}         & 2017 &                              & Score                &-  & -    & 82.5        & 88.6        \\
Zhao et al.         \cite{zhao2017two}             & 2017 &                               & Feature                & -    & -    & 83.7        & 93.7        \\

Baradel et al.      \cite{baradel2017pose}         & 2017 &                              & Score                 & 90.0 & -    & 84.8        & 90.6        \\
SI-MM               \cite{song2018skeleton}        & 2018 &                              & Feature                 & 91.9 & -    & 92.6        & 97.9        \\ 
Separable STA               \cite{das2019toyota}        & 2019 &                              & Feature                 & - & -    & 92.2        & 94.6        \\ 
SGM-Net               \cite{Jianan2020}        & 2020 &                              & Score                 & - & -    & 89.1        & 95.9        \\ 
VPN (RNX3D101)               \cite{das2020vpn}        & 2020 &                              & Feature                 & - & -    & 95.5        & 98.0        \\ 
\textcolor{black}{Luvizon et al.}              \cite{luvizon2020multi}        & 2020 &                              & Feature                 & - & -    & 89.9        & -        \\
\textcolor{black}{JOLO-GCN}              \cite{cai2021jolo}        & 2021 &                              & Score                 & - & -    & 93.8        & 98.1        \\
\textcolor{black}{TP-ViT}              \cite{jing2022tp}        & 2022 &                              & Feature,Score                 & - & -    & -        & -        \\
\textcolor{black}{RGBPose-Conv3D}              \cite{duan2022revisiting}        & 2022 &                              & Feature,Score                 & - & -    & 97.0        & 99.6        \\\hline
Rahmani et al.      \cite{rahmani2017learning}     & 2017 &            \multirow{4}{*}{S,D}                       & Feature               & -    & -    & 75.2        & 83.1        \\
Kamel et al.        \cite{kamel2018deep}           & 2018 &                                   & Score              & 88.1 & -    & -           & -           \\
3DSTCNN             \cite{zhao20193d}              & 2019 &                                  & Score              & -    & 95.3 & -           & -           \\
Rani et al.             \cite{rani2021kinematic}              & 2021 &                                   & Score             & -    & 88.5 & -           & -           \\\hline
DSSCA-SSLM          \cite{shahroudy2017deep}       & 2017 &             \multirow{5}{*}{RGB,S,D}              & Feature                & 97.5 & -    & 74.9        & -           \\
Deep Bilinear                \cite{hu2018deep}   & 2018 &                                  & Feature                & - & - & 85.4        & 90.7        \\ 
Cardenas et al.                \cite{cardenas2018multimodal}   & 2018 &                                 & Feature                & - & 94.6 & -        & -        \\
Khaire et al.                \cite{khaire2018combining}   & 2018 &                                 & Score                & - & 95.1 & -        & -        \\
Khaire et al.                \cite{khaire2018human}   & 2018 &                                 & Score               & - & 95.4 & -        & -        \\\hline
Ye et al.           \cite{ye2015temporal}          & 2015 & RGB,S,D,PC                      & Score              & -    & -    & -           & -           \\ 
Ardianto et al.           \cite{ardianto2018multi}          & 2018 & RGB,D,IR                       & Score             & -    & -    & -           & -           \\ 
FUSION-CPA          \cite{de2020infrared}          & 2020 & S,IR                              & Feature                & -    & -    & 91.6        & 94.5        \\
\textcolor{black}{ActAR}         \cite{lamghari2022actar}          & 2022 & S,IR                              & Feature                & -    & -    & -        & -        \\\hline
Wang et al.         \cite{wang2016exploring}       & 2016 & \multirow{11}{*}{RGB, Au}         & Feature,Score           & -    & -    & -           & -           \\
Owens et al.        \cite{owens2018audio}          & 2018 &                                   & Feature               & -    & -    & -           & -           \\
TBN                 \cite{kazakos2019epic}         & 2019 &                                   & Feature                & -    & -    & -           & -           \\
TSN$+$audio stream                 \cite{kazakos2019epic}         & 2019 &                                   & Score                & -    & -    & -           & -           \\
\textcolor{black}{AVSlowFast} \cite{xiao2020audiovisual}           & 2020 &                                  & Feature       & -          & -    & -    & -                      \\ 
Gao et al. \cite{gao2020listen}           & 2020 &                                  & Feature       & -          & -    & -    & -                      \\ 
\textcolor{black}{MAFnet} \cite{brousmiche2021multi}           & 2021 &                                  & Feature       & -          & -    & -    & -                      \\ 
\textcolor{black}{RNA-Net \cite{planamente2021cross}}           & 2021 &                                  & Feature,Score       & -          & -    & -    & -                      \\ 
\textcolor{black}{IMD-B \cite{alfasly2022learnable}}           & 2022 &                                  & Feature       & -          & -    & -    & -                      \\ 
\textcolor{black}{MM-ViT \cite{chen2022mm}}           & 2022 &                                  & Feature       & -          & -    & -    & -                      \\ 
\textcolor{black}{Zhang et al. \cite{zhang2022audio}}           & 2022 &                                  & Feature,Score       & -          & -    & -    & -                      \\ \hline
Dawar et al.        \cite{dawar2018data}         & 2018 & D,Ac,Gyr                      & Score                 & -    & 89.2    & -           & -           \\
Dawar et al.        \cite{dawar2018convolutional}         & 2018 & D,Ac,Gyr                      & Score                 & -    &-    & -           & -           \\
Wei et al.        \cite{wei2019fusion}         & 2019 & RGB,Ac,Gyr                      & Score                 & -    & 95.6    & -           & -           \\
Ahmad et al.          \cite{ahmad2019human}             & 2019 & D,Ac,Gyr                     & Feature,Score                 & -    &99.3    & -           & -           \\
MGAF          \cite{ahmad2020cnn}             & 2020 & D,Ac,Gyr                     & Feature                 & -    & 96.8    & -           & -           \\
\textcolor{black}{Pham et al.}          \cite{pham2021combining}             & 2021 & S,Ac                     & Feature                 & -    & 97.0    & -           & -           \\
\textcolor{black}{Ijaz et al.}          \cite{ijaz2022multimodal}             & 2022 & S,Ac                     & Feature,Score                 & -    & -    & -           & -           \\\hline
DCNN                \cite{jiang2015human}          & 2015 & Ac,Gyr                          & Feature                & -    & -    & -           & -           \\
DeepConvLSTM        \cite{ordonez2016deep}         & 2016 & Ac,Gyr                      & Feature                 & -    & -    & -           & -           \\
WiVi                \cite{zou2019wifi}             & 2019 & RGB,WiFi                          &  Score                  & -    & -    & -           & -           \\
Zou et al.          \cite{zou2020deep}             & 2020 & Ac,Gyr                      & Feature                 & -    & -    & -           & -           \\
Memmesheimer et al\cite{memmesheimer2020gimme}   & 2020 & S,WiFi,etc.                       & Feature                & -    & 93.3 & -           & -           \\
Imran et al         \cite{imran2020evaluating}     & 2020 & RGB,S,Gyr                    & Feature                 & -    & 97.9 & -           & -           \\ 
\textcolor{black}{Multi-Modal GCN \cite{shi2021multi}}     & 2021 & RGB,Audio,Text                    & Feature,Score                 & -    & - & -           & -           \\ 
\textcolor{black}{VATT \cite{akbari2021vatt}}     & 2021 & RGB,Audio,Text                    & Feature                 & -    & - & -           & -           \\ 
\textcolor{black}{Doughty et al. \cite{doughty2022you}}     & 2022 & RGB,Text                    & Feature                 & -    & - & -           & -           \\ 
\bottomrule[1pt]
\end{tabular}%
}
\label{5}
\end{table}

\subsection{FUSION}
As discussed in Section \ref{SINGLE MODALITY}, different modalities can have different strengths. 
Thus it becomes a natural choice to take advantage of the complementary strengths of different data modalities via fusion, so as to achieve enhanced HAR performance. 
There are two widely used multi-modality fusion schemes in HAR, namely, score fusion and feature fusion. Generally, the score fusion \cite{dhiman2020view} integrates the decisions that are separately made based on different modalities (e.g., by weighted averaging \cite{rani2021kinematic} or by learning a score fusion model \cite{wang2019generative}) to produce the final classification results. 
On the other hand, the feature fusion \cite{rahmani2017learning} generally combines the features from different modalities to yield aggregated features that are often very discriminative and powerful for HAR. 
Note that data fusion, i.e., fusing the multi-modality input data before feature extraction \cite{jiang2015human}, has also been exploited. Since the input data can be treated as the original raw features, here we simply categorize the data fusion approaches under feature fusion. 
Table \ref{5} gives the results of multi-modality fusion-based HAR methods on the MSRDailyActivity3D \cite{wang2012mining}, UTD-MHAD \cite{chen2015utd}, and NTU RGB+D \cite{shahroudy2016ntu} benchmark datasets. 

\subsubsection{Fusion of Visual Modalities}
With the emergence of low-cost RGB-D cameras, many multi-modality datasets \cite{wang2012mining, shahroudy2016ntu} have been created by the community, and consequently, some multi-modality fusion-based HAR methods have been proposed. Most of these methods focused on the fusion of visual modalities which are reviewed below. 

\noindent\textbf{Fusion of RGB and Depth Modalities.} 
The RGB and depth videos respectively capture rich appearance and 3D shape information, that are complementary and can be used for HAR.
Early methods \cite{kong2015bilinear, kong2017max} focused on extracting hand-crafted features which capture spatio-temporal structural relationships from the RGB and depth modalities for more reliable HAR. 
Since the current mainstream of research focuses on deep learning architectures, we review, in the following, deep learning methods which fuse RGB and depth data modalities. 
In \cite{imran2016human}, a four-stream deep CNN was introduced to extract features from different representations of depth data (i.e., three Depth Motion Maps \cite{yang2012recognizing} from three different viewpoints) and RGB data (i.e., a Motion History Image \cite{bobick2001recognition}). The output scores of these four streams were fused to perform action classification. 
By considering the depth and RGB modalities as a single entity, Wang et al. \cite{wang2017scene} extracted scene flow features from the spatially aligned and temporally synchronized RGB and depth frames. The bidirectional rank pooling \cite{fernando2016rank} was utilized to generate two dynamic images from the sequence of scene flow features. The dynamic images were then fed to two different CNNs, and finally, their classification scores were fused to perform HAR.
Wang et al. \cite{wang2018cooperative} represented the RGB and depth data as two pairs of RGB and depth dynamic images, which were then passed through a cooperatively trained CNN (c-ConvNet). The c-ConvNet consists of two streams that exploit features for action classification by jointly optimizing a ranking loss and a softmax loss. 
In \cite{wang2020hybrid}, a hybrid network that consists of multi-stream CNNs and 3D ConvLSTMs \cite{nips15shi} was introduced to extract features from RGB and depth videos. These features were then fused via canonical correlation analysis to perform action classification.
Wang et al. \cite{wang2019generative} proposed a generative framework to explore the feature distribution across the RGB and depth modalities. The fusion was performed by constructing a cross-modality discovery matrix, which was then fed to a modality correlation discovery network for the final prediction. 
Dhiman et al. \cite{dhiman2020view} designed a two-stream network composed of a motion stream and a Shape Temporal Dynamic (STD) stream to encode features from RGB and depth videos, respectively. Particularly, the motion stream takes a dynamic image from the RGB data as input and outputs classification scores. The STD network consists of the Human Pose Model in \cite{rahmani20163d} as its backbone, followed by several LSTMs and a softmax layer. The final classification scores were obtained by aggregating the scores of these two streams. 
Transformers have also become popular for multi-modal fusion \cite{li2021trear, zhou2022decoupling}. To handle the problem of tightly coupled spatio-temporal representation modeling, Zhou et al. \cite{zhou2022decoupling} proposed a decoupling and recoupling spatio-temporal representation approach, which disentangles learning spatio-temporal representations. Specifically, decoupled spatial and temporal networks are introduced to learn spatial and temporal independent discriminative features, respectively. A self-distillation-based re-coupling spatio-temporal network consisting of a multi-scale layer and Transformer blocks is then employed to model spatio-temporal dependencies. Finally, a cross-modal adaptive posterior fusion method is utilized to perform information aggregation. Li et al. \cite{li2021trear} introduced a Transformer framework for egocentric HAR, where RGB and its corresponding depth data are processed by an inter-frame attention encoder and then fused by a mutual attention block. 

\noindent\textbf{Fusion of RGB and Skeleton Modalities.} 
The appearance information provided by the RGB data, and the body posture and joint motion information provided by the skeleton sequences, are complementary and useful for activity analysis. Thus, several works have investigated deep learning architectures to fuse RGB and skeleton data for HAR. 
Zhao et al. \cite{zhao2017two} introduced a two-stream deep network, which consists of an RNN and a CNN to process skeleton and RGB data, respectively. Both the feature fusion and the score fusion were evaluated and the former achieved better performance. 
The two-stream architecture was also investigated in \cite{baradel2017human, baradel2017pose}, where the classification scores from a CNN model \cite{baradel2017pose} or an RNN model \cite{baradel2017human} trained on skeleton data and a skeleton conditioned spatio-temporal attention network trained on RGB data were fused for final classification. 
Zolfaghari \cite{zolfaghari2017chained} designed a three-stream 3D-CNN to process pose, motion, and the raw RGB images. The three streams were fused via a Markov chain model for action classification.
Liu et al. \cite{liu2017skeleton} proposed a spatio-temporal LSTM network, which is able to effectively fuse the RGB and skeleton features within the LSTM unit. 
In order to effectively capture the subtle variations in action performing, Song et al. \cite{song2018skeleton} proposed a learning framework containing two streams of skeleton-guided deep CNN to extract features from RGB and optical flow. Particularly, the local image patches around human body parts were first extracted from both RGB and flow sequences, and then fed into two individual CNNs followed by a part-aggregated pooling layer to generate two fixed-length feature vectors corresponding to RGB and flow frames. The skeleton data and the two sequences of RGB and flow feature vectors were then passed through a three-stream LSTM model followed by a fusion layer to obtain the final classification scores.
Das et al. \cite{das2019toyota} introduced a pose guided spatio-temporal attention network on top of a 3D CNN model taking RGB videos as input to perform HAR. 
In their subsequent work \cite{das2020vpn}, they extended their work in \cite{das2019toyota} by paying attention to the topology of the human body while computing the spatio-temporal attention maps. 
Li et al. \cite{Jianan2020} proposed a two-stream network, which consists of three main components namely the ST-GCN network \cite{yan2018spatial} to extract the skeleton features, R(2+1)D network \cite{tran2018closer} to extract RGB features, and a guided block which takes these features to enhance the action-related information in RGB videos. Finally, the score fusion approach was utilized to perform classification. 
\textcolor{black}{Cai et al. \cite{cai2021jolo} introduced a two-stream GCN network, respectively taking skeleton data and joint-aligned flow patches obtained from RGB videos as the inputs for HAR. }
\textcolor{black}{RGBPose-Conv3D \cite{duan2022revisiting} represents skeleton data as a 3D heatmap volume and passes the RGB video and the heatmap volume through a two-stream 3D CNN framework whose architecture is inspired by \cite{feichtenhofer2019slowfast}.}
Recently, Jing and Wang \cite{jing2022tp} proposed a Two-Pathway Vision Transformer (TP-ViT) to fuse RGB and skeleton data, which consists of a high-resolution path focusing on spatial information and a high-framerate path emphasizing the temporal information. 

\noindent\textbf{Fusion of Skeleton and Depth Modalities.} 
The skeleton data has been shown to be a succinct yet informative representation for human behavior analysis \cite{liu2019skeleton}, which, however, is a very sparse representation without the encoding of the shape information of the human body and the interacted objects.
Besides, skeleton data is often noisy, limiting the performance of HAR when it is used alone \cite{liu2017skeleton}. 
Meanwhile, depth maps provide discriminative 3D shape and silhouette information that can be helpful for HAR. Therefore, early methods \cite{rahmani2014real, shahroudy2015multimodal} have attempted to fuse hand-crafted features extracted from both the skeleton and the depth sequences for more accurate HAR. Recently, deep learning-based approaches have become the mainstream for fusing the skeleton and depth modalities. 
In order to learn the relationships between the human body and the objects, as well as the relations between different human body parts, Rahmani et al. \cite{rahmani2017learning} proposed an end-to-end learning framework, which consists of a CNN followed by the bilinear compact pooling and fully-connected layers for action classification. In particular, the model takes the relative geometry between every body part and others as the skeleton features and depth image patches around different body parts as the appearance features to encode body part-object and body part-body part relations for reliable HAR.
Kamel et al. \cite{kamel2018deep} utilized a three-stream CNN to extract action-related features from Depth Motion Images (DMIs), Moving Joint Descriptors (MJDs), and their combination. The DMI and MJD encode the depth and skeleton sequences as images. The output scores of these three CNNs were fused to obtain the final classification scores. 
Zhao et al. \cite{zhao20193d} utilized two 3D CNN streams taking raw depth data and Depth Motion Maps \cite{yang2012recognizing} as inputs, and a manifold representation stream taking 3D skeleton as input, for feature extraction from depth and skeleton sequences. The classification scores from these three networks were fused via score multiplications.
Rani et al. \cite{rani2021kinematic} proposed a three-stream 2D CNN to perform classification on three different hand-crafted features extracted from the depth and skeleton sequences, followed by a score fusion module to obtain the final classification result.

\noindent\textbf{Fusion of RGB, Skeleton, and Depth Modalities.}
Several hand-crafted feature-based methods \cite{hu2015jointly, kong2017max} have explored the fusion of RGB, skeleton, and depth modalities to further enhance the robustness of HAR. 
Meanwhile, deep learning-based methods \cite{shahroudy2017deep, hu2018deep, khaire2018combining, khaire2018human, cardenas2018multimodal} have received greater attention due to their superior performance. 
Shahroudy et al. \cite{shahroudy2017deep} focused on studying the correlation between modalities, and factorizing them into their correlated and independent components. Then a structured sparsity-based classifier was utilized for HAR. 
Hu et al. \cite{hu2018deep} learned the time-varying information across RGB, skeleton, and depth modalities by extracting temporal feature maps from each modality, and then concatenating them along the modality dimension. These multi-modal temporal features were then fed to a stack of bilinear blocks to exploit the mutual information from both modality and temporal directions. 
Khaire et al. \cite{khaire2018combining} proposed a five-stream CNN network, which takes Motion History Images \cite{bobick2001recognition}, Depth Motion Maps \cite{yang2012recognizing} (i.e., from three different viewpoints), and skeleton images generated respectively from RGB, depth, and skeleton sequences as inputs. Each CNN was trained individually, and the output scores of these five streams were fused with a weighted product model \cite{triantaphyllou1998multi} to obtain the final classification scores. 
In their subsequent work \cite{khaire2018human}, three fusion methods were explored for combining skeletal, RGB, and depth modalities. 
Cardenas and Chavez \cite{cardenas2018multimodal} fed three different optical spectra channels from skeleton data \cite{hou2016skeleton} and dynamic images from both RGB and depth videos to a pre-trained CNN to extract multi-modal features. Finally, a feature aggregation module was used to perform classification. 


\noindent\textbf{Other Fusion Methods.}
Besides, some other multi-modality fusion methods were proposed \cite{ardianto2018multi, de2020infrared}. 
In the work of \cite{ardianto2018multi}, a multi-modal fusion model was built on top of the TSN framework \cite{wang2016temporal}. The RGB, depth, infrared, and optical flow sequences were passed through TSNs to perform initial classification, followed by a fusion network applied on the initial classification scores to obtain the final classification scores. 
Main and Noumeir \cite{de2020infrared} utilized a 3D CNN and a 2D CNN to extract features from infrared and skeleton data, respectively. The feature vectors were then concatenated and exploited by a multi-layer perceptron for HAR. 

\subsubsection{Fusion of Visual and Non-visual Modalities}
Visual and non-visual modalities can also be fused to leverage their complementary discriminative capabilities for more accurate and robust HAR models. 

\noindent\textbf{Fusion of Audio and Visual Modalities.}
Audio data provides complementary information to the appearance and motion information in the visual data. Several deep learning-based methods have been proposed to fuse these two types of modalities for HAR. 
Wang et al. \cite{wang2016exploring} introduced a three-stream CNN to extract multi-modal features from audio signal, RGB frames, and optical flows. Both feature fusion and score fusion were evaluated, and the former achieved better performance.
Owens and Efros \cite{owens2018audio} trained a two-stream CNN in a self-supervised manner to identify any misalignment between the audio and visual sequences. The learned model was then fine-tuned on the HAR datasets for audio-visual HAR. 
Inspired by TSN \cite{wang2016temporal}, Kazakos et al. \cite{kazakos2019epic} introduced a Temporal Binding Network (TBN), which takes audio, RGB, and optical flow as inputs for egocentric HAR. The TBN network contains a three-stream CNN to fuse the multi-modal input sequences within each Temporal Binding Window, followed by a temporal aggregation module for classification. Their results showed that the TBN outperformed the Temporal Segment Network (TSN) \cite{wang2016temporal} for the audio-visual HAR task. 
Gao et al. \cite{gao2020listen} utilized audio signal to reduce temporal redundancies in videos. Particularly, this method distills the knowledge from a teacher network trained on video clips to a student network trained on image-audio pairs for efficient HAR. The student network is a two-stream deep model which takes the starting frame and audio signal of the video clip as input followed by a feature fusion, fully-connected, and softmax layers for audio-visual HAR. 
\textcolor{black}{
Inspired by the work of \cite{feichtenhofer2019slowfast}, Xiao et al. \cite{xiao2020audiovisual} introduced a hierarchically integrated audio-visual representation framework containing slow and fast visual pathways that are deeply integrated with a faster audio pathway at multiple layers. Two training strategies, including randomly dropping the audio pathway and hierarchical audio-visual synchronization, were also introduced to enable joint training of audio and video modalities. 
}

Recently, Alfasly et al. \cite{alfasly2022learnable} adopted a transformer model, i,e., BERT~\cite{devlin2018bert}, to obtain the sentence-based semantic embeddings of each textual label in audio and video datasets. A Semantic Audio-Video Label Dictionary (SAVLD) is built in which each video label is mapped into $k$ audio labels of which they all are considered semantically similar. Then, an Irrelevant Modality Dropout (IMD) module was proposed to select the most relevant audio modality embedding for each video. Finally, both video modality embedding and the selected audio modality embedding are fused for action classification. 
Zhang et al. \cite{zhang2022audio} proposed an audio-adaptive model leveraging the rich audio information to adjust the visual representation, along with an audio-infused recognizer to maintain domain-irrelevant features. Likewise, Planamente et al. \cite{planamente2021cross} proposed a novel Relative Norm Alignment (RNA) loss to align audio and visual modalities for egocentric HAR. 
Besides, Chen and Ho \cite{chen2022mm} presented a Multi-Modal Video Transformer named MM-ViT for compressed video action recognition by factorizing the self-attention mechanism of ViT~\cite{dosovitskiy2020image} over the space, time and modality dimensions.

\noindent\textbf{Fusion of Acceleration and Visual Modalities.}
Some hand-crafted feature-based methods \cite{zou2017robust, elmadany2018multimodal} have exploited the fusion of acceleration and visual modalities for HAR. 
Besides, many deep learning methods \cite{dawar2018convolutional, dawar2018data,  wei2019fusion, ahmad2019human, ahmad2020cnn,pham2021combining} have also been proposed recently. 
Dawar et al. \cite{dawar2018convolutional} represented inertial signal as an image, and utilized two CNNs to fuse the depth images and inertial signals using score fusion. 
Due to the limited depth-inertial training data, Dawar et al. \cite{dawar2018data} proposed a data augmentation framework based on depth and inertial modalities, which were separately fed to a CNN and a CNN+LSTM respectively. The scores of the two models were fused during testing for better classification. 
Wei et al. \cite{wei2019fusion} respectively fed the 3D video frames and 2D inertial images to a 3D CNN and a 2D CNN for HAR, and the score fusion achieved better performance than feature fusion. 
Several depth-inertial fusion techniques have also been investigated in \cite{ahmad2019human, ahmad2020cnn} by designing different two-stream CNN architectures, where the inertial signal was transformed into images using the technique in \cite{jiang2015human}. 
Recently, Ijaz et al. \cite{ijaz2022multimodal} proposed a multi-modal Transformer for nursing action recognition, where the correlations between skeleton and acceleration data are modeled by transformers.


\noindent\textbf{Other Fusion Methods.}
Memmesheimer et al. \cite{memmesheimer2020gimme} transformed the skeleton, inertial, and WiFi data to a color image, which was fed to a CNN to perform HAR. 
Imran and Raman \cite{imran2020evaluating} designed a three-stream architecture, where a 1D CNN, a 2D CNN, and an RNN were used for gyroscopic, RGB, and skeleton data, respectively. Several fusion methods were adopted to predict the final class label, while fusing features using canonical correlation analysis achieved the best performance. 
Zou et al. \cite{zou2019wifi} designed a two-stream architecture, named WiVi, which takes the RGB and WiFi frames as inputs of the C3D and CNN streams. A multi-modal score fusion module was built on top of the network to perform HAR. 
Recently, Doughty and Snoek \cite{doughty2022you} leveraged adverb pseudo-labels in RGB videos for semi-supervised fine-grained HAR. 
Shi et al.\cite{shi2021multi} proposed Multi-modal GCNs to explore modality-aware multi-action relations from RGB, audio, and text data. Akbari et al. \cite{akbari2021vatt} presented a Video-Audio-Text Transformer (VATT) taking the linear projection of video, audio, and text data as the input to extract multi-modal representations. VATT quantifies the granularity of the different modalities, and is trained by adopting the Noise Contrastive Estimation (NCE) that aligns video-audio pairs as well as video-text pairs.


Although the aforementioned multi-modality fusion-based HAR methods have achieved promising results on some benchmark datasets, the task of effective modality fusion is still largely open. Specifically, most of the existing multi-modality methods have complicated architectures which require high computational costs. Thus efficient multi-modality HAR also needs to be addressed. 

\subsection{CO-LEARNING}
Co-learning explores how knowledge learned from auxiliary modalities can be used to assist the learning of a model on another modality \cite{baltruvsaitis2018multimodal}. 
Transferring knowledge among different modalities can overcome the shortcomings of a single data modality and enhance its performance.
Unlike fusion methods, in co-learning methods, the data of the auxiliary modalities is only required during training rather than testing. This is particularly beneficial in cases where some modalities are missing during testing. 
Co-learning can also benefit the learning of a certain modality with fewer samples, by leveraging other correlated modalities with richer samples for assisting model training. 



\subsubsection{Co-Learning with Visual Modalities} 
Most of the co-learning-based HAR methods focused on co-learning with visual modalities, such as RGB with depth modalities, and RGB with skeleton modalities. 

\noindent\textbf{Co-Learning with RGB and Depth Modalities.}
Knowledge transfer \cite{baltruvsaitis2018multimodal} between the RGB modality which captures appearance information, and depth data which encodes 3D shape information, has been shown to be useful for improving the representation capability of each modality for HAR, especially when one of these modalities has a limited amount of annotated data for training. 
There have been some methods that used hand-crafted features for co-learning \cite{kong2015bilinear, kong2017max}. 
Recently, Garcia et al. \cite{garcia2018modality, garcia2019dmcl, garcia2019learning} proposed several deep learning-based HAR methods based on cross-modality knowledge distillation. 
In \cite{garcia2018modality}, a knowledge distillation framework was proposed to distill knowledge from a teacher network taking depth videos as input to a RGB-based student network. The knowledge distillation was achieved by forcing the feature maps and prediction scores of the student network to be similar to the teacher network. 
While in \cite{garcia2019learning}, an adversarial learning-based knowledge distillation strategy was proposed to learn the student stream. 
In their subsequent work \cite{garcia2019dmcl}, a three-stream network taking RGB, depth, and optical flow data as inputs, was trained using a cooperative learning strategy, i.e., the predicted labels generated by the modality stream with the minimum classification loss were used as an additional supervision for the training of other streams. 

\noindent\textbf{Co-Learning with RGB and Skeleton Modalities.} 
Co-learning between the RGB and skeleton modalities has also been investigated in existing works \cite{mahasseni2016regularizing, thoker2019cross, song2020modality}. 
Mahasseni and Todorovic \cite{mahasseni2016regularizing} 
utilized a CNN+LSTM network to perform classification based on RGB videos, and an LSTM model trained on skeleton data to act as a regularizer by forcing the output features of the two models (i.e., CNN+LSTM and LSTM) to be similar. 
Thoker and Gall \cite{thoker2019cross} utilized TSN \cite{wang2016temporal} as the teacher network taking RGB videos, and an ensemble of student networks (e.g., ST-GCN \cite{yan2018spatial} and HCN \cite{li2018co}) handling skeleton data, to perform knowledge distillation for cross-modality HAR. Specifically, each student network was trained from the supervisory information provided by the teacher network as well as other student networks. 
In \cite{song2020modality}, a supervision signal from an auxiliary LSTM network taking the skeleton data as input was fed to a two-stream CNN+LSTM architecture taking RGB and optical flow sequences as inputs. The auxiliary supervision signal forces the model to align the distributions of the auxiliary (skeleton) and source (RGB and optical flow) modalities, which thus enhances the representation of the RGB and optical flow modalities. 
Hong et al. \cite{hong2021video} exploited the Video Pose Distillation (VPD) strategy in a teacher-student architecture, where the teacher network, i.e., an off-the-shelf pose estimator, provides a weak supervision for training the student network extracting robust pose representation from RGB videos. 


\noindent\textbf{Other Co-Learning Methods.} 
Besides using RGB with depth or RGB with skeleton data for co-learning, other visual modalities have also been investigated. 
For example, Wang et al. \cite{wang2018pm} learned a transferable generative model which takes an infrared video as input and generates a fake feature representation of its corresponding RGB video. The discriminator consists of two sub-networks, i.e., a deep binary classifier which attempts to differentiate between the generated fake features and the real RGB feature representation, and a predictor which takes both the infrared video representation and the generated features as inputs to perform action classification. 
Chadha et al. \cite{chadha2019neuromorphic} embedded the PIX2NVS emulator \cite{bi2017pix2nvs} (i.e., converting pixel domain video frames to neuromorphic events) into a teacher-student framework, which transfers knowledge from a pre-trained optical flow teacher network to a neuromorphic event student network. 


\subsubsection{Co-Learning with Visual and Non-visual Modalities}
There are also several works on co-learning between visual and non-visual modalities \cite{kong2019mmact, liu2021semantics, perez2020audio}.
Kong et al. \cite{kong2019mmact} 
and Liu et al. \cite{liu2021semantics} trained multiple teacher networks on non-visual modalities, such as acceleration, gyroscope, and orientation signal, to teach an RGB video-based student network through knowledge distillation with an attention mechanism. 
In \cite{kong2019mmact}, the knowledge was transferred by first integrating the classification scores of the teachers with attention weights learned by feeding the concatenated classification scores of the teachers to a feed-forward neural network. Then the attended-fusion scores were used as an additional supervision to train the student network. 
In \cite{liu2021semantics}, 
knowledge distillation was achieved by forcing the visual explanations (i.e., attention maps that highlight the important regions for the classification scores) of both the student and teacher networks to be similar to bridge the modality gap. 
In another work, Perez et al. \cite{perez2020audio} proposed a knowledge distillation framework to transfer knowledge from a teacher network trained on RGB videos to a student network taking the raw sound data as input. Specifically, an objective function \cite{Hinton2015distil}, which encourages the student model to predict the true labels as well as matching the soft labels provided by the teacher network, was utilized. 
Recently, Yang et al. \cite{yang2022interact} proposed a Cross-modal Interactive Alignment (CIA) model for unsupervised domain adaptive video action recognition. Specifically, CIA consists of a Mutual Complementarity (MC) module to refine the transferability of each modality and a Spatial Consensus (SC) module to achieve cross-modal consensus.

Besides fusing multiple modalities and transferring knowledge between different modalities, there are also a few works that leverage the correlation between different modalities for self-supervised learning. 
For example, Alwassel et al. \cite{alwassel2019self} leveraged unsupervised clustering in the audio/video modality as the supervisory signal for the video/audio modality, respectively. 
The audio-visual correspondence and temporal synchronization have also been used as self-supervised signals to learn discriminative audio and visual features \cite{arandjelovic2017look, korbar2018cooperative}. 
\textcolor{black}{
Several other works \cite{miech2019howto100m, miech2020end} leverage narrations of videos as a weak supervisory signal to jointly learn video and text representations for action recognition and detection. 
Miech et al. \cite{miech2019howto100m} introduced a text-video embedding model, which is trained from the caption-clip pairs using the max-margin ranking loss for learning joint representations of video and language automatically. 
In their further study \cite{miech2020end}, a novel training loss derived from Multiple Instance Learning and Noise Contrastive Estimation, was introduced to address misalignments in narrated videos. 
Besides, Sun et al. \cite{sun2019videobert} introduced a joint visual-linguistic model by extending BERT \cite{devlin2018bert} to videos for zero-shot HAR.}
Likewise, Lin et al. \cite{lin2022cross} proposed a cross-modal Transformer architecture leveraging videos and text labels for zero-shot HAR. 
Specifically, a ResNet-Transformer integrates the learning of visual representations and visual-semantic associations into a unified architecture. At inference, the model only takes a new single-modality input (unseen action video) as input to produce its visual representation for zero-shot HAR. 


\section{DATASETS} \label{DATASET}
A large number of datasets have been created to train and evaluate the HAR methods. Table \ref{6} lists a series of benchmark datasets. In particular, the attributes of these datasets are also summarized. 

For RGB-based HAR, the UCF101 \cite{soomro2012ucf101}, HMDB51 \cite{kuehne2011hmdb}, and Kinectis-400 \cite{kay2017kinetics} are widely used as benchmark datasets. Besides, Kinetics-600 \cite{carreira2018short}, Kinetics-700 \cite{carreira2019short}, EPIC-KITCHENS-55 \cite{damen2018scaling}, THUMOS Challenge 15 \cite{THUMOS15}, ActivityNet \cite{caba2015activitynet}, and Something-Something-v1 \cite{goyal2017something} datasets are also popularly used. 
For 3D skeleton, depth, infrared, and point cloud-based HAR, the large-scale NTU RGB+D \cite{shahroudy2016ntu} and NTU RGB+D 120 \cite{liu2019ntu} are widely used benchmark datasets. Besides, the MSRDailyActivity3D \cite{wang2012mining}, Northwestern-UCLA \cite{wang2014cross}, and UWA3D Multiview II \cite{rahmani2016histogram} datasets are also widely used for depth-based HAR, while the
InfAR \cite{gao2016infar} dataset is often used for infrared-based HAR. 
Some point cloud-based methods \cite{Wang_2021_WACV, liu2019meteornet} have also been evaluated on point cloud data obtained from depth images. 
DvsGesture \cite{amir2017low} and DHP19 \cite{calabrese2019dhp19} datasets are popular datasets for event stream-based HAR. 
However, there are no widely used benchmark datasets for non-visual modality-based HAR. 

In addition, for multi-modality-based HAR,
the NTU RGB+D \cite{shahroudy2016ntu}, NTU RGB+D 120 \cite{liu2019ntu}, MMAct \cite{kong2019mmact}, and EPIC-KITCHENS \cite{damen2018scaling, damen2022rescaling} are large benchmark datasets suitable for fusion and co-learning of different modalities. The MSRDailyActivity3D \cite{wang2012mining}, UTD-MHAD \cite{chen2015utd}, and PKU-MMD \cite{liu2017pku} datasets are also popularly used.

\begin{table}[t]
\caption{\footnotesize Some representative benchmark datasets with various data modalities for HAR. S: Skeleton, D: Depth, IR: Infrared, PC: Point Cloud, ES: Event Stream, Au: Audio, Ac: Acceleration, Gyr: Gyroscope, EMG: Electromyography.}
\vspace{-3mm}
\centering
\resizebox{0.47\textwidth}{!}{%
\setlength{\tabcolsep}{1.0pt}
\begin{tabular}{lcccccc}
\toprule[1pt]
\textbf{Dataset} &
  \textbf{Year} &
  \textbf{Modality} &
  \textbf{\#Class} &
  \textbf{\#Subject} &
  \textbf{\#Sample} &
  \textbf{\#Viewpoint} \\ \hline
KTH                 \cite{schuldt2004recognizing}   & 2004 & RGB                     & 6     & 25  & 2,391       & 1   \\ 
Weizmann            \cite{gorelick2007actions}      & 2005 & RGB                     & 10    & 9   & 90        & 1   \\ 
IXMAS               \cite{weinland2006free}         & 2006 & RGB                     & 11    & 10  & 330       & 5   \\ 
HDM05               \cite{muller2007documentation}  & 2007 & RGB,S                   & 130   & 5   & 2,337      & 1   \\ 
Hollywood           \cite{laptev2008learning}       & 2008 & RGB                     & 8      & -         & 430   & -  \\ 
Hollywood2          \cite{marszalek2009actions}     & 2009 & RGB                     & 12    & -  & 3,669     & -   \\ 
MSR-Action3D        \cite{li2010action}             & 2010 & S,D                     & 20    & 10  & 567       & 1   \\ 
Olympic             \cite{niebles2010modeling}      & 2010 & RGB                     & 16    & -   & 783       & -   \\ 
CAD-60              \cite{sung2011human}            & 2011 & RGB,S,D                 & 12    & 4   & 60        & -   \\ 
HMDB51              \cite{kuehne2011hmdb}           & 2011 & RGB                     & 51    & -   & 6,766     & -   \\ 
RGB-HuDaAct         \cite{ni2011rgbd}               & 2011 & RGB,D                   & 13    & 30  & 1,189     & 1   \\ 
ACT4$^2$            \cite{cheng2012human}           & 2012 & RGB,D                   & 14    & 24  & 6,844     & 4   \\ 
DHA                 \cite{lin2012human}             & 2012 & RGB,D                   & 17    & 21  & 357       & 1   \\ 
MSRDailyActivity3D \cite{wang2012mining}           & 2012 & RGB,S,D                 & 16    & 10  & 320       & 1   \\ 
UCF101              \cite{soomro2012ucf101}         & 2012 & RGB                     & 101   & -   & 13,320    & -   \\ 
UTKinect            \cite{xia2012view}              & 2012 & RGB,S,D                 & 10    & 10  & 200       & 1   \\ 
Berkeley MHAD       \cite{ofli2013berkeley}         & 2013 & RGB,S,D,Au,Ac          & 12    & 12  & 660       & 4   \\ 
CAD-120             \cite{koppula2013learning}      & 2013 & RGB,S,D                 & 10    & 4   & 120       & -   \\ 
IAS-lab             \cite{munaro20133d}             & 2013 & RGB,S,D,PC              & 15    & 12  & 540       & 1   \\ 
J-HMDB              \cite{jhuang2013towards}        & 2013 & RGB,S                     & 21    & -   & 31,838    & -   \\ 
MSRAction-Pair      \cite{oreifej2013hon4d}         & 2013 & RGB,S,D                 & 12    & 10  & 360       & 1   \\ 
UCFKinect           \cite{ellis2013exploring}       & 2013 & S                       & 16    & 16  & 1,280     & 1   \\ 
Multi-View TJU      \cite{liu2014multiple}          & 2014 & RGB,S,D                 & 20    & 22  & 7,040     & 2   \\ 
Northwestern-UCLA   \cite{wang2014cross}            & 2014 & RGB,S,D                 & 10    & 10  & 1,475     & 3   \\ 
Sports-1M           \cite{karpathy2014large}        & 2014 & RGB                     & 487   & -   & 1,113,158 & -   \\ 
UPCV                \cite{theodorakopoulos2014pose} & 2014 & S                       & 10    & 20  & 400       & 1   \\ 
UWA3D Multiview \cite{rahmani2014hopc, rahmani2014action} &
  2014 &
  RGB,S,D &
  30 &
  10 &
  $\sim$900 &
  4 \\ 
ActivityNet         \cite{caba2015activitynet}      & 2015 & RGB                     & 203   & -   & 27,801    & -   \\ 
SYSU 3D HOI                \cite{hu2015jointly}            & 2015 & RGB,S,D                 & 12    & 40  & 480       & 1   \\ 
THUMOS Challenge 15                \cite{THUMOS15}            & 2015 & RGB                 & 101    & -  & 24,017       & -   \\ 
TJU                 \cite{liu2015coupled}           & 2015 & RGB,S,D                 & 15    & 20  & 1,200     & 1   \\ 
UTD-MHAD            \cite{chen2015utd}              & 2015 & RGB,S,D,Ac,Gyr         & 27    & 8   & 861       & 1   \\ 
UWA3D Multiview II \cite{rahmani2016histogram} & 2015 & RGB,S,D & 30 & 10 & 1,075 & 4 \\ 
\textcolor{black}{Charades}               \cite{sigurdsson2016hollywood}             & 2016 & RGB                      & 157     & 267  &9,848        &-    \\ 
InfAR               \cite{gao2016infar}             & 2016 & IR                      & 12    & 40  & 600       & 2   \\ 
NTU RGB+D           \cite{shahroudy2016ntu}         & 2016 & RGB,S,D,IR              & 60    & 40  & 56,880    & 80  \\ 
YouTube-8M          \cite{abu2016youtube}           & 2016 & RGB                     & 4,800 & -   & 8,264,650 & -   \\ 
AVA                 \cite{gu2018ava}                & 2017 & RGB                     & 80    & -   & 437    & -   \\ 
DvsGesture \cite{amir2017low}                      & 2017 & ES                     & 17    & 29   & -    & -   \\ 
FCVID               \cite{jiang2017exploiting}      & 2017 & RGB                     & 239   & -   & 91,233    & -   \\ 
Kinetics-400            \cite{kay2017kinetics}          & 2017 & RGB                     & 400   & -   & 306,245  & -   \\ 
NEU-UB              \cite{kong2017max}              & 2017 & RGB,D                   & 6     & 20  & 600       & -   \\ 
PKU-MMD             \cite{liu2017pku}               & 2017 & RGB,S,D,IR              & 51    & 66  & 1,076     & 3   \\ 
Something-Something-v1 \cite{goyal2017something} &
  2017 &
  RGB &
  174 &
  - &
  108,499 &
  - \\ 
UniMiB SHAR         \cite{micucci2017unimib}        & 2017 & Ac                      & 17    & 30  & 11,771    & -   \\ 
EPIC-KITCHENS-55                 \cite{damen2018scaling}  & 2018 & RGB,Au    & -     & 32  & 39,594         & Egocentric   \\ 
Kinetics-600            \cite{carreira2018short}          & 2018 & RGB                     & 600   & -   & 495,547   & -   \\ 
RGB-D Varying-view  \cite{ji2018large}              & 2018 & RGB,S,D                 & 40    & 118 & 25,600     & 8+1(360$^\circ$)   \\ 
DHP19          \cite{calabrese2019dhp19}          & 2019 & ES, S              & 33    & 17  & -         & 4   \\ 
Drive\&Act          \cite{martin2019drive}          & 2019 & RGB,S,D,IR              & 83    & 15  & -         & 6   \\ 
Hernang{\'o}mez et al. \cite{hernangomez2019human} &2019 &Radar &8 & 11 &1,056 &-\\
Kinetics-700            \cite{carreira2019short}          & 2019 & RGB                     & 700   & -   & 650,317   & -   \\ 
Kitchen20           \cite{moreaux2019benchmark}     & 2019 & Au                      & 20    & -   & 800       & -   \\ 
MMAct               \cite{kong2019mmact}            & 2019 & RGB,S,Ac,Gyr,etc. & 37    & 20  & 36,764    & 4+Egocentric   \\ 
Moments in Time     \cite{monfort2019moments}       & 2019 & RGB                     & 339   & -   & $\sim$1,000,000 & -   \\ 
Wang et al. \cite{wang2019joint} &2019 & WiFi CSI & 6 & 1 &1,394 &-\\
NTU RGB+D 120       \cite{liu2019ntu}               & 2019 & RGB,S,D,IR              & 120   & 106 & 114,480   & 155 \\ 
ETRI-Activity3D \cite{jang2020etri} & 2020 & RGB,S,D              & 55   & 100 & 112,620   & - \\
EV-Action           \cite{wang2020ev}               & 2020 & RGB,S,D,EMG             & 20    & 70  & 7,000     & 9   \\ 
IKEA ASM \cite{ben2020ikea} & 2020 & RGB,S,D              & 33   & 48 & 16,764    & 3 \\
RareAct \cite{miech2020rareact} & 2020 & RGB              & 122   &-  &905    & - \\
\textcolor{black}{BABEL} \cite{punnakkal2021babel} & 2021 & Mocap              & 252   &-  &13,220    & - \\
\textcolor{black}{HAA500} \cite{chung2021haa500} & 2021 & RGB              & 500   &-  &10,000    & - \\
\textcolor{black}{HOMAGE} \cite{rai2021home} & 2021 & RGB,IR,Ac,Gyr,etc.              & 75   &27  &1,752    & 2$\sim$5 \\
\textcolor{black}{MultiSports} \cite{li2021multisports} & 2021 & RGB              & 66   &-  &37,701    & - \\
\textcolor{black}{UAV-Human} \cite{li2021uav} & 2021 & RGB,S,D,IR,etc.              & 155   & 119  & 67,428    & - \\
\textcolor{black}{Ego4D} \cite{grauman2022ego4d} & 2022 & RGB,Au,Ac,etc.              & -   & 923 & -   & Egocentric \\
EPIC-KITCHENS-100 \cite{damen2022rescaling} & 2022 & RGB,Au,Ac              & -   & 45 & 89,979   & Egocentric \\
\textcolor{black}{JRDB-Act} \cite{ehsanpour2022jrdb} & 2022 & RGB,PC              & 26   & -  & 3,625    & 360$^\circ$ \\
\bottomrule[1pt]
\end{tabular}%
}
\label{6}
\end{table}

\section{DISCUSSION} \label{DISCUSSION}
In the previous sections, we review the methods and datasets for HAR with various data modalities. Below we discuss some of the potential and important directions that could need further investigation in this domain. 

\textbf{Datasets.}
Large and comprehensive datasets generally have a vital importance for the development of HAR, especially for the deep learning-based HAR methods. 
There are many aspects which indicate the quality of a dataset, such as its size, diversity, applicability, and type of modality.
Despite the large number of existing datasets that have advanced the HAR area greatly, to further facilitate the research on HAR, new benchmark datasets 
are still required. For example, most of the existing multi-modality datasets were collected in controlled environments, where actions were usually performed by volunteers. Thus collecting multi-modality data from uncontrolled environments to achieve large and challenging benchmarks, for further promoting multi-modality HAR in practical applications, can be important. Besides, the construction of large and challenging datasets for every person’s action recognition in crowded environments can also be further investigated. 
Group action recognition~\cite{han2022dual, perez2022skeleton} and human-human interaction recognition \cite{ke2016human, perez2019interaction} tasks also demand diverse datasets. 


\textbf{Multi-modality Learning.}
As discussed in Section \ref{MULTI-MODALITY}, 
several multi-modality learning methods, including multi-modality fusion and cross-modality transfer learning, have been proposed for HAR. The fusion of multi-modality data which can often complement each other, results in improvements of HAR performance, while co-learning can be used to handle the issue of the lack of data of some modalities. 
However, as pointed out by \cite{wang2020makes},
many existing multi-modality methods are not as effective as expected owing to a series of challenges, such as over-fitting. This implies there remain opportunities to design more effective fusion and co-learning strategies for multi-modality HAR.  


\textbf{Efficient Action Analysis.}
The superior performance of many HAR methods is built on high computational complexity, while efficient HAR is also crucial for many real-life practical applications. 
Hence how to reduce the computational costs and resource consumption (e.g., CPU, GPU, and energy consumption), and achieve efficient and fast HAR, deserve further studies \cite{lin2019tsm, cheng2020skeleton}.

\textbf{Early Action Recognition.}
\textcolor{black}{Early action recognition (action prediction) enables recognition when only a part of the action has been performed, i.e., recognizing an action before it has been fully performed \cite{li2020hard, liu2018ssnet, fernando2021anticipating, wang2021dear}.}
This is also an important problem due to its relevance in some applications, such as online human-robot interaction and early alarm in some real-life scenarios. 


\textbf{Few-shot Action Analysis.}
It can be difficult to collect a large amount of training data (especially multi-modality data) for all action classes. 
To handle this issue, one of the possible solutions is to take advantage of few-shot learning techniques \cite{wang2020generalizing, xie2021few}. 
Though there have been some attempts for few-shot HAR \cite{liu2019ntu, zhang2020few}, considering the significance of handling the issues of scarcity of data in many practical scenarios, more advanced few-shot action analysis can still be further explored. 


\textbf{Unsupervised and Semi-supervised Learning.}
Supervised learning methods, especially deep learning-based ones, often require a large amount of data with expensive labels for model training. 
\textcolor{black}{Meanwhile, unsupervised and semi-supervised learning techniques \cite{alayrac2016unsupervised, singh2021semi, song2021spatio, yang2021skeleton} often enable to leverage the availability of unlabelled data to train the models, which significantly reduces the requirement of large labeled datasets. }
Since unlabelled action samples are often easier to be collected than labeled ones, unsupervised and semi-supervised HAR are also an important research direction that is worthy of further development. 


\section{CONCLUSION}
HAR is an important task that has attracted significant research attention in the past decades, and various data modalities with different characteristics have been used for this task. 
In this paper, we have given a comprehensive review of HAR methods using different data modalities. 
Besides, multi-modality recognition methods, including fusion and co-learning methods, have been also surveyed. 
Benchmark datasets have also been reviewed, and some potential research directions have been discussed. 

\bibliographystyle{ieeetr.bst}

\bibliography{review}

\end{document}